\documentclass[journal]{IEEEtran}

\usepackage{cite}
\usepackage{amsmath,amssymb,amsfonts}
\usepackage{graphicx}
\usepackage{booktabs}

\usepackage{multirow}
\usepackage{stfloats}
\usepackage{adjustbox}
\usepackage{enumitem}
\usepackage{hyperref}
\usepackage{array}
\usepackage{url}
\usepackage{subcaption}

\usepackage[table]{xcolor}
\usepackage{xcolor}

\newcommand{\Dmono}{\textbf{D}^{\text{mono}}}
\newcommand{\Dmvs}{\textbf{D}^{\text{mvs}}}
\newcommand{\Gmvs}{\textbf{G}^{\text{mvs}}}
\newcommand{\CVmulti}{\textbf{CV}_{\text{multi}}}
\newcommand{\CVmono}{\textbf{CV}_{\text{mono}}}
\newcommand{\PV}{\textbf{PV}}
\newcommand{\PVcurve}{\overline{\text{\textbf{PV}}}}
\newcommand{\xmvs}{\textbf{x}_k^{\text{mvs}}}
\newcommand{\xmono}{\textbf{x}_k^{\text{mono}}}
\newcommand{\DmvsStep}{\textbf{D}_k^{\text{mvs}}}
\newcommand{\DmonoStep}{\textbf{D}_k^{\text{mono}}}
\newcommand{\CmvsStep}{\textbf{C}_k^{\text{mvs}}}
\newcommand{\CmvsHat}{{\hat{\text{\textbf{C}}}}_k^{\text{mvs}}}
\newcommand{\CmvsNext}{\textbf{C}_{k+1}^{\text{mvs}}}
\newcommand{\Encg}{\text{Enc}_g}
\newcommand{\Encmvs}{\text{Enc}_{\text{mvs}}}
\newcommand{\Encmono}{\text{Enc}_{\text{mono}}}
\newcommand{\Enccross}{\text{Enc}_{\text{cross}}}
\newcommand{\hmono}{\textbf{h}_k^{\text{mono}}}
\newcommand{\hmonoPrev}{\textbf{h}_{k-1}^{\text{mono}}}
\newcommand{\hmonoCand}{\hat{\text{\textbf{h}}}_k^{\text{mono}}}
\newcommand{\hmvs}{\textbf{h}_k^{\text{mvs}}}
\newcommand{\DeltaDmono}{\Delta \textbf{D}_k^{\text{mono}}}
\newcommand{\DeltaDmvs}{\Delta \textbf{D}_k^{\text{mvs}}}
\newcommand{\FFPN}{\textbf{F}_{\text{FPN}}^l}

\begin{document}

\title{M2Depth: Unifying Monocular Depth Foundation Priors with Multi-View Stereo}
\author{Byeonggwon Lee, Sanggi Lee, Siwoo Lee, Khang Truong Giang, and Soohwan Song%
\thanks{Byeonggwon Lee, Sanggi Lee, Siwoo Lee, and Soohwan Song are with the Department of Computer Science and Artificial Intelligence, Dongguk University, Seoul 04620, Republic of Korea.}%
\thanks{Khang Truong Giang is with 42dot, Seongnam-si, Gyeonggi-do 13449, Republic of Korea.}%
}

\IEEEaftertitletext{%
\vspace{-0.8\baselineskip}
\begin{center}
\includegraphics[width=0.95\textwidth]{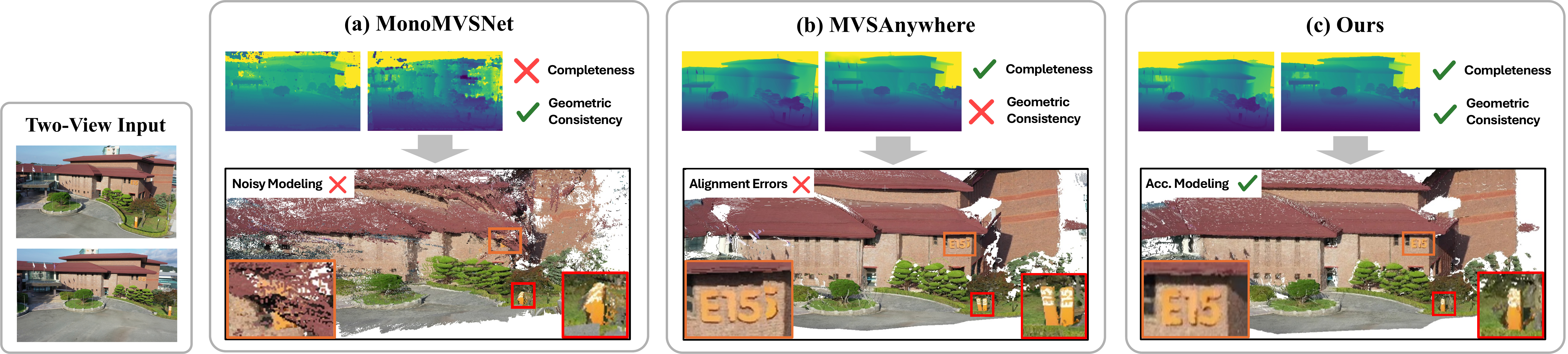}
\end{center}
\vspace{0.4em}
\refstepcounter{figure}\label{fig:overview}%
\footnotesize\noindent\textbf{Fig.~\thefigure.} 3D reconstruction comparison of DFM-based MVS methods in a sparse-view setting. (a) The MVS-centric method (MonoMVSNet \cite{jiang2025monomvsnet}) fails to produce a clean or generalizable reconstruction. (b) The DFM-centric method (MVSAnywhere \cite{izquierdo2025mvsanywhere}) generates a clean depth map but lacks geometric consistency, suffering from scale ambiguity and alignment errors. (c) In contrast, our method produces both a clean depth map and an accurate, geometrically aligned 3D structure.
\vspace{0.8\baselineskip}
}

\maketitle

\begin{center}
\footnotesize
This work has been submitted to the IEEE for possible publication.
Copyright may be transferred without notice, after which this version
may no longer be accessible.
\end{center}

\begin{abstract}
Deep learning-based Multi-View Stereo (MVS) has advanced significantly but often generalizes poorly to unseen scenes, particularly in occluded areas or regions with limited view overlap. To mitigate this, recent approaches integrate Depth Foundation Models (DFMs) into MVS pipelines to provide monocular depth priors. However, existing methods typically rely on a static, one-way fusion scheme, which fails to fully exploit the complementary strengths of both modalities. We propose a novel framework that overcomes this limitation by tightly coupling a DFM with a cascade MVS pipeline through a bidirectional mutual refinement strategy. Our method leverages MVS depth to resolve the scale ambiguity in monocular predictions, while the monocular depth, in turn, enhances the structural completeness and fine-grained detail of the MVS estimate. Furthermore, we introduce a prior-guided cost volume refinement mechanism that effectively integrates multi-view and monocular information via attention-based fusion and discretized depth bins, thereby promoting local geometric consistency.
Extensive experiments demonstrate that our method outperforms state-of-the-art MVS approaches on standard benchmarks, producing more complete and generalizable depth maps with sharp boundaries. Furthermore, although not explicitly designed for sparse-view settings, our framework generalizes remarkably well, competing favorably with even dedicated sparse-view methods while maintaining a superior accuracy-efficiency trade-off.
%Extensive experiments demonstrate that our approach outperforms state-of-the-art MVS methods on benchmark datasets, producing highly complete and generalizable depth maps with sharp object boundaries, even in sparse-view cases.
\end{abstract}


\section{Introduction}
\label{sec:intro}

Multi-View Stereo (MVS) aims to reconstruct detailed 3D geometries by estimating dense pixel correspondences across images captured from multiple viewpoints. Recent deep learning-based approaches \cite{yao2018mvsnet, gu2020cascade, ma2024confident, shin2023mosaicmvs, son2024cmvde} generate robust features and utilize end-to-end optimization frameworks, significantly outperforming traditional MVS methods \cite{schonberger2016pixelwise, galliani2015massively}. Despite these advances, MVS approaches often struggle to generalize well to unseen scene structures, particularly under complex geometries, viewpoint variations, and challenging lighting conditions.
Additionally, accuracy deteriorates in regions with occlusions or limited view overlap, where depth estimation becomes challenging due to ambiguous scene geometry.

Concurrently, Vision Foundation Models (VFMs) \cite{awais2025foundation, oquab2023dinov2} have shown impressive generalization capabilities across diverse visual tasks. Among these, Depth Foundation Models (DFMs), such as DepthAnythingV2 \cite{yang2024depth2}, trained on extensive image-depth datasets, yield robust monocular depth predictions even in previously unseen domains, generating sharp and clear depth maps from single images. However, their monocular nature leads to scale ambiguity and a lack of explicit multi-view consistency, thereby limiting their direct application to accurate 3D reconstruction.

To overcome these limitations, two primary approaches have recently been proposed to integrate DFMs \cite{yang2024depth2} with MVS. The first approach (e.g., MonoMVSNet \cite{jiang2025monomvsnet}) is an MVS-centric method that incorporates a monocular prior into an MVS pipeline. This prior provides structural guidance in depth boundary regions, which improves depth estimation accuracy. The second approach (e.g., MVSAnywhere \cite{izquierdo2025mvsanywhere}) employs a DFM-centric pipeline that leverages multi-view information as a prior. It combines multi-view cost volumes with a pre-trained DFM encoder to generate more complete depth maps.

Despite these performance gains, both lines of work share a fundamental limitation: they treat the complementary modality as a static, one-way prior, fusing it without a feedback mechanism. This prevents each stream from dynamically compensating for the other’s inherent limitations. Consequently, MVS-centric pipelines still struggle to produce clean, generalizable depth in textureless or occluded regions, while DFM-centric pipelines remain prone to scale ambiguity, yielding misaligned geometry. Moreover, this unidirectional fusion causes errors from an inaccurate prior to propagate without correction through subsequent stages.

In this paper, we propose a novel framework that tightly couples a DFM with a cascade MVS pipeline \cite{cao2024mvsformer++}, moving beyond the conventional, static one-way fusion paradigm \cite{jiang2025monomvsnet, izquierdo2025mvsanywhere}. Our approach introduces a co-refinement strategy, ensuring that both monocular and MVS depth representations are progressively improved throughout the cascade. This method operates iteratively and bidirectionally, where the monocular and MVS depths mutually reference and complement one another. Specifically, the MVS depth is leveraged to improve the local scale consistency of the monocular depth, while the monocular depth provides a structural prior to enhance the details and completeness of the MVS depth.

Additionally, we propose a cost volume refinement mechanism that effectively fuses multi-view and monocular information. Our method constructs a monocular cost volume and integrates it with the multi-view cost volume via an attention-based mechanism. We also discretize the monocular depth into depth bins that act as structural priors, promoting local consistency within the fused volume. Consequently, our approach generates highly complete depth maps, particularly in challenging regions with occlusions or limited view overlap. As illustrated in Fig.~\ref{fig:overview}, unlike existing methods \cite{jiang2025monomvsnet, izquierdo2025mvsanywhere}, our method produces both clean depth maps and an accurate, geometrically aligned 3D structure, demonstrating its robust integration of DFM and MVS strengths.

Our main contributions can be summarized as follows:
\begin{itemize}[leftmargin=0.5cm]
    \item We introduce a novel MVS framework that integrates monocular priors from DFMs, enabling clean and complete depth maps in challenging cases while ensuring accurate 3D reconstruction.
    \item We propose a bidirectional mutual refinement mechanism that leverages MVS depth to align monocular scale and, in turn, uses monocular depth to enhance MVS structure.
    \item We propose a prior depth-guided cost volume refinement strategy, fusing monocular and multi-view volumes using bin masks to improve spatial consistency.
    \item Our method achieves high accuracy and strong generalization across multiple benchmarks \cite{aanaes2016large, Knapitsch2017, schroeppel2022robust}, surpassing existing MVS methods in both depth estimation and 3D reconstruction. It also generalizes well to sparse-view settings, outperforming even dedicated sparse-view methods. % Source code for our method is publicly available.\footnote{https://github.com/released-after-acceptance}    
\end{itemize}

\section{Related Work}
\subsection{Multi-View Stereo}
% Traditional MVS methods \cite{galliani2015massively, schonberger2016pixelwise} typically rely on handcrafted features for stereo matching, often struggling with untextured regions and repetitive patterns. To address these limitations, 
Recent deep learning-based approaches \cite{yao2018mvsnet} have significantly improved reconstruction accuracy. MVSNet \cite{yao2018mvsnet}, for example, introduced an end-to-end pipeline consisting of feature extraction, cost volume construction, and cost volume regularization. Despite achieving high accuracy, MVSNet uses computationally expensive 3D convolutions, leading to substantial GPU memory requirements and limited scalability.

Subsequent works proposed cascade cost-volume frameworks \cite{gu2020cascade} to alleviate this issue. These methods progressively refine depth hypotheses across multiple stages, reducing computational complexity by narrowing the search space at each stage. Consequently, these cascade architectures have been widely adopted as a standard baseline, inspiring numerous extended MVS approaches \cite{liao2022wt, zhang2023geomvsnet, wu2024gomvs, ma2024confident, shin2023mosaicmvs} in recent years. %For instance, local and global attention mechanisms \cite{liao2022wt} have been introduced to effectively refine feature representations.
Geometry-aware methods \cite{zhang2023geomvsnet, wu2024gomvs} explicitly incorporate structural information and surface normals to improve precision and consistency in depth estimation. Furthermore, Transformer-based approaches, such as MVSFormer \cite{cao2022mvsformer}, have emerged to enhance feature extraction capabilities. Its enhanced version, MVSFormer++ \cite{cao2024mvsformer++}, introduces side-view attention, refined attention scaling, and advanced positional encoding to reinforce inter-view fusion and consistency.

\begin{figure*}[t]
    \centering
    \includegraphics[width=0.90\textwidth]{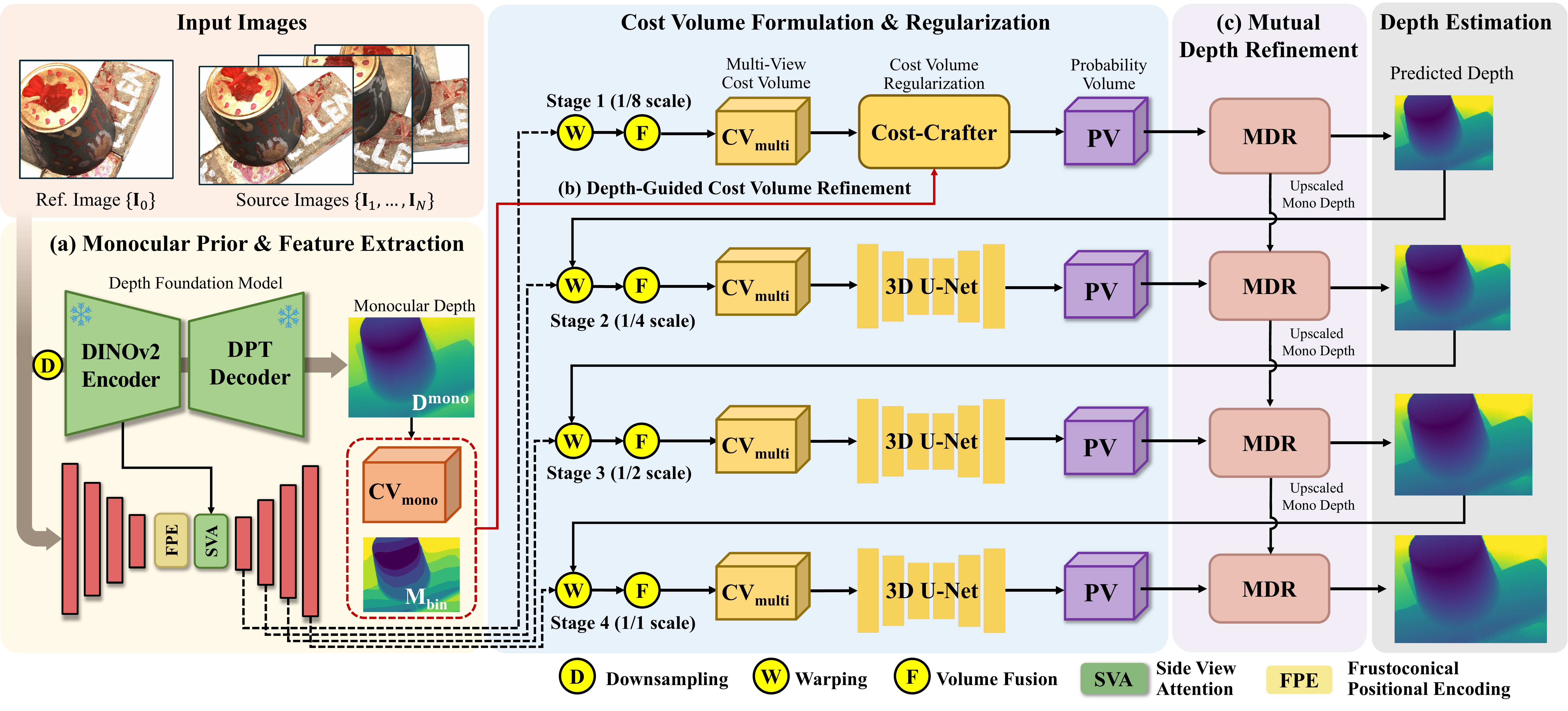}
    \caption{Our system follows a cascaded framework. Similar to the baseline~\cite{cao2024mvsformer++}, we construct and regularize a multi-view cost volume $\textbf{CV}_{\text{multi}}$ at each stage, which is then converted into a probability volume $\textbf{PV}$. The MVS depth map $\textbf{D}^{\text{mvs}}$ is then estimated from $\textbf{PV}$ by selecting the depth hypothesis with the highest probability. Unlike the baseline, our method (a) extracts a monocular depth map $\textbf{D}^{\text{mono}}$ using the DFM~\cite{yang2024depth2}, (b) refines the cost volume in the first stage using $\textbf{D}^{\text{mono}}$, and (c) performs iterative mutual refinement between the MVS depth and the monocular depth at all stages.
}
    \label{fig:pipeline}
\end{figure*}

\subsection{Depth Foundation Model}
Vision Foundation Models (VFMs) \cite{awais2025foundation, oquab2023dinov2} are models trained on extensive datasets, providing strong generalization to various visual tasks with little to no fine-tuning.
In monocular depth estimation, Depth Foundation Models (DFMs) have recently gained attention.
%Depth Pro \cite{bochkovskii2024depth} provides zero-shot monocular depth predictions characterized by exceptional sharpness and detail.
Dense Prediction Transformer \cite{ranftl2021vision} leverages transformer architectures to enhance dense prediction tasks with improved structural and contextual understanding. DepthAnything \cite{yang2024depth2} further exploits web-scale image-depth datasets to enable robust, zero-shot depth estimation across diverse scenes.
However, these monocular models inherently suffer from scale ambiguity, hindering their direct application in 3D modeling.

Recent research \cite{zhou2025all, wen2025foundationstereo, jiang2025defom} has incorporated DFMs into stereo depth estimation frameworks, exploiting their strong generalization and semantic priors. MonSter \cite{cheng2025monster} and FoundationStereo \cite{wen2025foundationstereo} propose frameworks to effectively fuse monocular depth priors into the stereo matching pipeline.
%StereoAnywhere \cite{bartolomei2025stereo} presents a hybrid stereo-monocular method leveraging DFM priors to improve depth predictions in challenging regions, such as textureless or occluded areas.
These works underscore the potential of DFMs to enhance stereo matching and generalization performance.

While DFM integration is common in stereo vision \cite{zhou2025all, wen2025foundationstereo, jiang2025defom}, fewer studies apply it to MVS \cite{jiang2025monomvsnet, izquierdo2025mvsanywhere}. Relatedly, Murre~\cite{guo2025multi} leverages SfM-guided monocular depth for multi-view reconstruction, and Marigold-DC~\cite{viola2025marigold} guides a monocular diffusion model using sparse metric depth. For MVS specifically, MonoMVSNet \cite{jiang2025monomvsnet} takes an MVS-centric route, injecting a static monocular prior into its cost volume. However, this static nature risks propagating initial depth errors. Alternatively, MVSAnywhere \cite{izquierdo2025mvsanywhere} uses a DFM-centric pipeline, conditioning a monocular model on multi-view cost volumes; yet, its loose coupling with multi-view geometry often results in misaligned reconstructions.

In contrast, our method uses a bidirectional co-refinement strategy where MVS and monocular estimates mutually correct each other iteratively. This tight coupling resolves DFM scale ambiguity while enhancing MVS completeness, avoiding error propagation and geometric misalignment.

\subsection{Sparse-View Reconstruction}

Sparse-view reconstruction aims to recover 3D geometry from few images, a setting where conventional MVS often fails due to ambiguous matching. Recent methods like VolRecon~\cite{ren2023volrecon}, UFORecon~\cite{na2024uforecon}, and SparseRecon~\cite{han2025sparserecon} tackle this using neural implicit or volume-rendering formulations.
%Additionally, FatesGS~\cite{huang2025fatesgs} and GSRecon~\cite{yang2025gsrecon} employ Gaussian splatting to balance efficiency and quality.
However, these methods are highly specialized for sparse-view settings, often relying on computationally expensive test-time optimization or implicit rendering. Consequently, they diverge from the explicit multi-view depth estimation and cost-volume reasoning of standard MVS pipelines~\cite{cao2022mvsformer, cao2024mvsformer++, jiang2025rrt}, failing to explore the joint refinement of monocular priors and multi-view geometry within a general framework.

In contrast, our DFM-guided MVS framework is rooted in explicit multi-view geometry rather than specialized sparse-view pipelines~\cite{ren2023volrecon, na2024uforecon, han2025sparserecon}. Avoiding expensive implicit surface optimization, we enhance a cascade MVS~\cite{cao2024mvsformer++} via prior-guided cost construction and bidirectional mutual depth refinement. This iterative correction is crucial: monocular priors compensate for missing correspondences, while MVS geometry mitigates scale ambiguity and hallucinations. Consequently, our method generalizes to sparse views without fine-tuning or test-time refinement. This superior accuracy-efficiency trade-off proves that tightly coupling monocular priors with MVS reasoning enables robust sparse-view reconstruction.

\section{Method}
Our objective is to estimate a high-quality depth map $\textbf{D}_0$ for a reference image 
$\textbf{I}_0 \in \mathbb{R}^{3 \times H \times W}$, given $N$ source images 
$\{ \textbf{I}_i \}_{i=1}^N$ and their corresponding camera parameters. 
As illustrated in Fig.~\ref{fig:pipeline}, our framework integrates a Depth Foundation Model (DFM), 
specifically DepthAnythingV2~\cite{yang2024depth2}, into a cascade MVS pipeline~\cite{gu2020cascade}. 

The core MVS pipeline employs a four-stage coarse-to-fine approach, 
progressively refining depth estimates from $1/8$ to full resolution. 
To achieve this, it extracts multi-scale visual features 
$\{\textbf{F}_{\text{FPN}}^l \}_{l=1}^4 \in \mathbb{R}^{C_l \times H_l \times W_l}$ 
using a feature pyramid network (FPN)~\cite{lin2017feature}. 
At each stage $l$, stereo matching is performed using the corresponding features $\textbf{F}_{\text{FPN}}^l$. The multi-view cost volume $\mathbf{CV}_{\text{multi}}$ is built by measuring photometric or feature differences between the reference feature and the source features warped under all depth hypotheses. To refine this raw volume, most prior methods \cite{cao2022mvsformer, cao2024mvsformer++, wu2024gomvs} employ a 3D U-Net \cite{yao2018mvsnet}, producing a more discriminative representation. It then applies softmax along depth to obtain the probability volume $\mathbf{PV}$, from which the MVS depth $\mathbf{D}^{\text{mvs}}$ is estimated by selecting the depth hypothesis with the highest probability.

Concurrently, the DFM estimates a monocular depth prior 
$\Dmono \in \mathbb{R}^{H_1 \times W_1}$ 
for the reference image $\textbf{I}_0$, where $H_1 = \tfrac{H}{8}$ and $W_1 = \tfrac{W}{8}$
. 
This predicted depth map $\Dmono$ provides structural priors 
and fine-grained shape cues. Therefore, it is used to guide the depth refinement process at each cascade stage, enhancing the completeness 
of the final estimation.

To effectively leverage this prior depth information, we introduce two key contributions:

\begin{enumerate}
  \item \textbf{Prior-Guided 
  Cost Volume Refinement} (Sect.~III-A): Applied at the initial cascade stage, this mechanism enhances the multi-view cost volume. It integrates a monocular cost volume (derived from $\Dmono$) and structural bin masks to establish a robust geometric foundation for the subsequent stages.
  \item \textbf{Mutual Depth Refinement} (Sect.~III-B): This module operates across all stages, performing mutually complementary updates. It simultaneously refines the MVS depth $\Dmvs$ to enhance fine-grained structural details, while also using the multi-view geometric cues to achieve pixel-accurate scale alignment for the monocular depth $\Dmono$.
\end{enumerate}

%By integrating monocular priors, which provide global structure, with MVS features that offer geometric precision, our method generates highly accurate and detailed depth maps. Unlike prior approaches \cite{izquierdo2025mvsanywhere, jiang2025monomvsnet}, we jointly upsample and refine the MVS and monocular depths at each stage. % As shown in Fig. ~\ref{fig:fig3}, this iterative co-refinement process progressively enhances the scale consistency of the monocular prior, ensuring robust performance even when the initial priors from the DFM are significantly mis-scaled or noisy. 

% \begin{figure}[t]
% \centering
% \includegraphics[width=0.90\linewidth]{sec/Figure/figure3-1.pdf}
% \caption{Depth map results at Stage 1 and 4. (a) Reference image. MonoMVSNet \cite{jiang2025monomvsnet} may produce a noisy Stage 1 depth map (b), and these artifacts propagate to its final Stage 4 result (c). While our initial Stage 1 depth (d) is similarly noisy, our mutual refinement module (e) effectively cleans the depth map, enhancing object boundaries and surface consistency. This provides a robust foundation, leading to a much cleaner final depth map (f).}
% \label{fig:fig3} 
% \end{figure}

\subsection{Prior Depth-Guided Cost Volume Refinement}
\label{sec:cost_crafter}

We propose a prior depth-guided cost volume refinement module, termed \textit{Cost-Crafter}, which fuses a conventional multi-view cost volume with a monocular prior-derived cost volume to regularize ambiguous matching regions using global depth structure.
This refinement is applied only at the first cascade stage, where reliable global geometry is crucial for establishing a strong coarse representation and the computational overhead of global reasoning remains affordable.
%This refinement is applied only at the first cascade stage, where reliable global geometry is most important for establishing a strong coarse representation and where the computational overhead of global reasoning remains affordable.

We first construct the multi-view cost volume $\CVmulti \in \mathbb{R}^{C\times D\times H_1\times W_1}$ by aggregating stereo matching costs across multi-view features, where $C$ and $D$ denote the channel and depth dimensions, respectively. In parallel, we build a monocular cost volume $\CVmono \in \mathbb{R}^{1\times D\times H_1\times W_1}$ from the prior depth map using soft one-hot encoding. To provide a compact structural prior, the continuous monocular depth is discretized into $M$ uniformly spaced bins, producing a depth-bin mask $\textbf{M}_{\text{bin}} \in \{1,\ldots,M\}^{H_1\times W_1}$. These three components, $\CVmulti$, $\CVmono$, and $\textbf{M}_{\text{bin}}$, are used as inputs to Cost-Crafter.

\begin{figure}[t]
\centering
\includegraphics[width=1.0\linewidth]{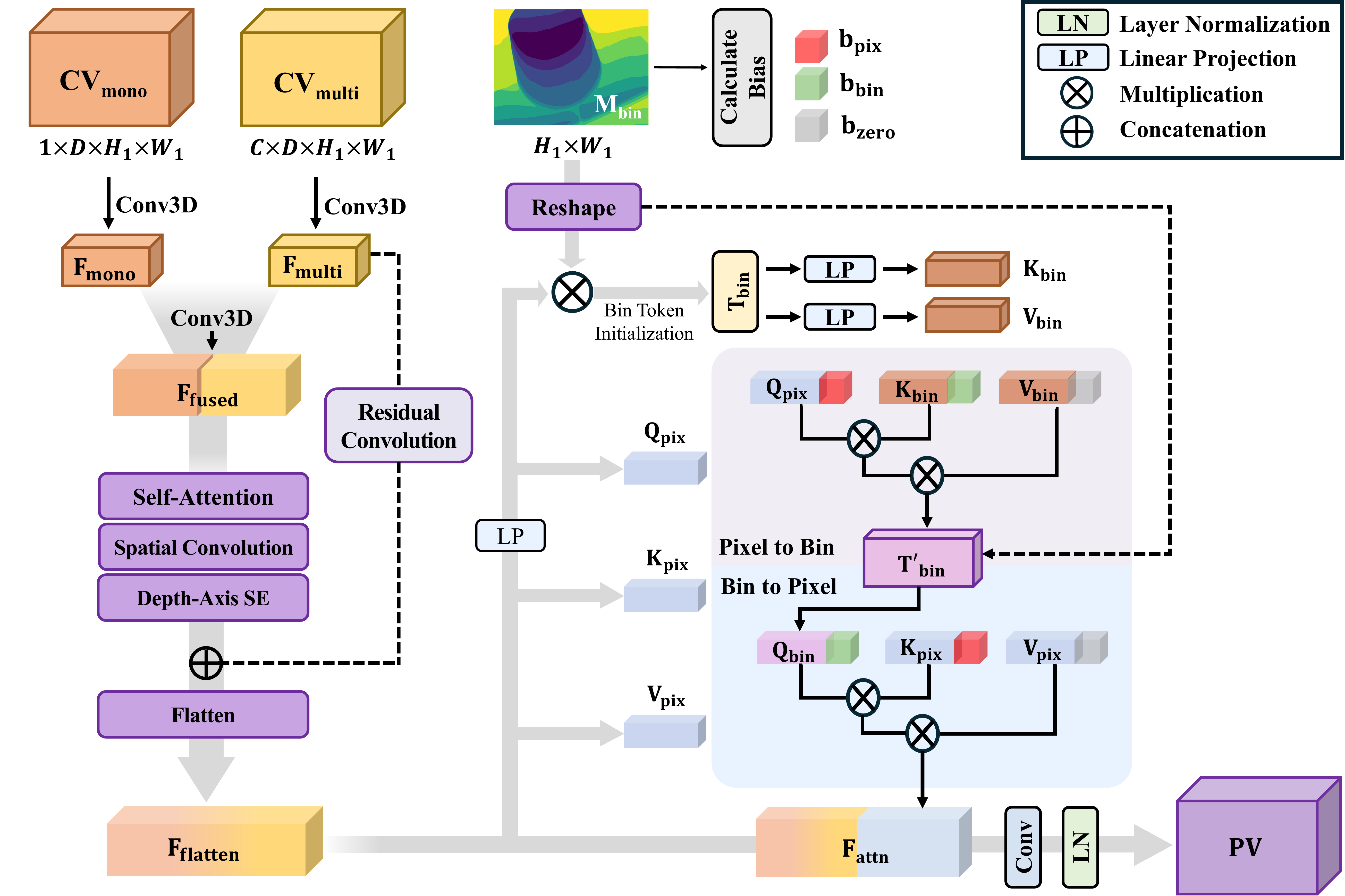}
\caption{Architecture of the Cost-Crafter. This module enhances the multi-view cost volume by incorporating a monocular cost volume and structural bin masks, which serve as spatial priors.}
\label{fig:fig4}
\end{figure}

% \noindent\textbf{Cost-Crafter module.}
As illustrated in Fig.~\ref{fig:fig4}, $\CVmulti$ and $\CVmono$ are first projected by separate 3D convolutions to obtain intermediate features $\textbf{F}_{\text{multi}}$ and $\textbf{F}_{\text{mono}}$. The two features are concatenated along the channel dimension and fused by another 3D convolution, yielding the fused volume $\textbf{F}_{\text{fused}} \in \mathbb{R}^{C\times D\times H_1\times W_1}$.
This fused representation combines photometric evidence from stereo matching with structural cues from the monocular prior.

To perform global reasoning, we reshape the fused volume into pixel-level features $\textbf{F}_{\text{flatten}} \in \mathbb{R}^{S\times C}$, where $S=D\times H_1\times W_1$, by flattening depth and spatial dimensions. Direct global attention over $S$ locations is expensive and lacks depth order. We therefore introduce a compact bin-based tokenization strategy guided by prior depth bins.

\noindent\textbf{Depth-Bin Tokenization.}
Using $\textbf{M}_{\text{bin}}$, we group pixels that share similar prior depth levels. Specifically, the bin mask is replicated along the depth dimension and flattened into $\textbf{M}_{\text{flatten}} \in \{1,\ldots,M\}^{S}$, then converted to a one-hot matrix $\textbf{M}_{\text{bin}}^{\text{one-hot}} \in \{0,1\}^{S\times M}$.
Representative bin tokens $\textbf{T}_{\text{bin}} \in \mathbb{R}^{M\times C}$ are computed by masked average pooling:
\[
\textbf{T}_{\text{bin}}
=
\frac{\left(\textbf{M}_{\text{bin}}^{\text{one-hot}}\right)^T \textbf{F}_{\text{flatten}}}
{\text{counts}_{\text{bin}}+\epsilon},
\]
where $\text{counts}_{\text{bin}} \in \mathbb{R}^{M\times 1}$ denotes the number of assigned pixels in each bin and $\epsilon$ is a small constant for numerical stability. This aggregation compresses the dense pixel space into only $M$ representative depth tokens, where typically $M \ll S$, enabling efficient global context modeling while preserving depth-aware structure.

\noindent\textbf{Ordinal Depth Encoding via Scalar Bias.}
Standard cross-attention is permutation-invariant and therefore does not explicitly model the ordinal relationship between different depth layers. To inject this geometric structure, we introduce scalar positional bias terms for pixels and bins:
\begin{align*}
\textbf{b}_{\text{pix}} &= \textbf{M}_{\text{flatten}}\sqrt{2\mu_{\text{rel}}} \in \mathbb{R}^{S}, \\
\textbf{b}_{\text{bin}} &= [1,2,\ldots,M]\sqrt{2\mu_{\text{rel}}} \in \mathbb{R}^{M},
\end{align*}
where $\mu_{\text{rel}}$ is a learnable scalar controlling the bias strength. Instead of explicitly forming an additional bias matrix, we augment the query ($\textbf{Q}$), key ($\textbf{K}$), value ($\textbf{V}$) features with these scalar terms:
\begin{align*}
\tilde{\textbf{Q}} = [\text{LP}_Q(\textbf{Q}),\, \textbf{b}_q], \quad
\tilde{\textbf{K}} = [\text{LP}_K(\textbf{K}),\, \textbf{b}_k], \quad
\tilde{\textbf{V}} = [\text{LP}_V(\textbf{V}),\, \textbf{0}],
\end{align*}
where $\text{LP}_Q$, $\text{LP}_K$ and $\text{LP}_V$ are separate learned linear projections for queries, keys, and values.
We then define the shared cross-attention as
\[
\text{Attn}(\textbf{Q},\textbf{K},\textbf{V})
=
\left[
\text{softmax}\left(\frac{\tilde{\textbf{Q}}\tilde{\textbf{K}}^T}{\sqrt{C+1}}\right)\tilde{\textbf{V}}
\right]_{:,1:C}.
\]
Since $\tilde{\textbf{Q}}\tilde{\textbf{K}}^T = \textbf{Q}\textbf{K}^T + \textbf{b}_q \textbf{b}_k^T$, the added scalar term acts as an ordinal depth prior that favors interactions between features at similar depth levels. Importantly, we slice the output $[\cdot]_{:, 1:C}$ to ensure that the bias terms influence only the attention scores (weights) and do not pollute the semantic features propagated through the Value ($\mathbf{V}$) path.

\noindent\textbf{Bidirectional Pixel-Bin Interaction.}
The module performs two passes of cross-attention to propagate information between pixels and depth bins. In the first pass (\textit{pixel-to-bin}), pixel features query the bin tokens:
\[
\mathbf{F}_{\text{pix}\rightarrow\text{bin}}
=
\text{Attn}\big(
\text{LP}_Q(\mathbf{F}_{\text{flatten}}),\,
\text{LP}_K(\mathbf{T}_{\text{bin}}),\,
\text{LP}_V(\mathbf{T}_{\text{bin}})
\big).
\]
This step allows each pixel, including those in textureless, occluded, or weakly matched regions, to aggregate global context from structurally similar depth bins. The bin tokens are then updated by re-aggregating the attended pixel responses:
\[
\mathbf{T}'_{\text{bin}}
=
\text{LayerNorm}
\left(
\mathbf{T}_{\text{bin}}
+
\frac{\left(\mathbf{M}_{\text{bin}}^{\text{one-hot}}\right)^T \mathbf{F}_{\text{pix}\rightarrow\text{bin}}}
{\text{counts}_{\text{bin}}+\epsilon}
\right).
\]
This update injects fine-grained pixel evidence back into the compact bin representation.

In the second pass (\textit{bin-to-pixel}), the updated bin tokens query the pixel features:
\[
\mathbf{F}_{\text{bin}\rightarrow\text{pix}}
=
\text{Attn}\big(
\text{LP}_Q(\mathbf{T}'_{\text{bin}}),\,
\text{LP}_K(\mathbf{F}_{\text{flatten}}),\,
\text{LP}_V(\mathbf{F}_{\text{flatten}})
\big).
\]
Let $\mathbf{W}_{\text{bin}\rightarrow\text{pix}}$ denote the attention weights of this step. We broadcast the refined global responses back to the original pixel space using the transposed weights:
\[
\tilde{\mathbf{F}}_{\text{bin}\rightarrow\text{pix}}
=
\mathbf{W}_{\text{bin}\rightarrow\text{pix}}^T
\mathbf{F}_{\text{bin}\rightarrow\text{pix}}.
\]
The final attention-enhanced feature is obtained through a residual connection and normalization:
\[
\mathbf{F}_{\text{attn}}
=
\text{LayerNorm}
\left(
\mathbf{F}_{\text{flatten}}
+
\tilde{\mathbf{F}}_{\text{bin}\rightarrow\text{pix}}
\right)
\in \mathbb{R}^{S\times C}.
\]
This bidirectional design first aggregates reliable global depth cues into compact bin tokens and then redistributes the refined information back to ambiguous pixels, improving structural coherence while preserving local detail.

Finally, $\mathbf{F}_{\text{attn}}$ is reshaped back to the 3D volume form and processed by a $1\times1\times1$ convolution, 3D LayerNorm, and softmax along the depth dimension to produce the refined probability volume $\mathbf{PV} \in [0,1]^{D\times H_1\times W_1}$.
By integrating monocular depth priors, bin-level global reasoning, and ordinal depth-aware attention, Cost-Crafter regularizes the initial cost volume toward more complete and geometrically consistent depth estimation, especially in regions where conventional photometric matching is unreliable.

\begin{figure}[t]
\centering
\includegraphics[width=0.90\linewidth]{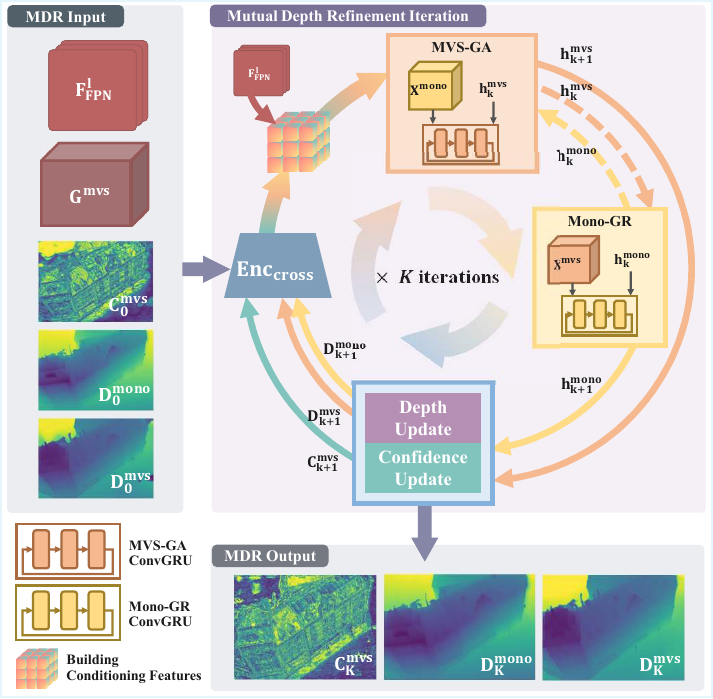}
\caption{Architecture of the mutual depth refinement module. The module employs two symmetric recurrent (GRU) branches that iteratively update the MVS and monocular depth maps along with the MVS confidence map.}
\label{fig:fig5}
\end{figure}

\subsection{Mutual Depth Refinement}
\label{sec:mdr}

At each cascade stage, we perform \emph{Mutual Depth Refinement} (MDR) between the monocular depth $\Dmono$ and the MVS depth $\Dmvs$. This process serves a dual purpose: correcting the spatially varying scale ambiguity of $\Dmono$ at the pixel level, and enhancing the structural completeness and local consistency of $\Dmvs$.
Unlike prior MVS-fusion methods \cite{jiang2025monomvsnet} using a single global scale-shift correction, our approach iteratively predicts pixel-wise residuals. This ensures locally accurate scale alignment while preserving geometric details.

To achieve this, we extend mutual refinement strategies, originally developed for pairwise stereo matching \cite{cheng2025monster, wen2025foundationstereo}, to the cascade MVS framework. However, adapting these strategies to MVS requires handling complex challenges, such as multi-view coupling, cross-view uncertainty aggregation, and coarse-to-fine depth evolution. To specifically address these MVS-specific requirements, we introduce tailored components: geometry-volume conditioning, a curvature-based confidence cue, and an adaptive anchoring design.

As illustrated in Fig.~\ref{fig:fig5}, MDR consists of two symmetric branches:
\textbf{MVS-Guided Alignment} (MVS-GA) and \textbf{Mono-Guided Refinement} (Mono-GR).
The two branches operate iteratively with ConvGRU \cite{cho2014learning} blocks that predict residual updates for the opposite modality.
This bidirectional design yields conservative yet effective refinement: reliable MVS geometry anchors the monocular depth, while the monocular prior restores structure where photometric matching is weak.
% This bidirectional design yields conservative but effective refinement: reliable MVS geometry anchors the monocular depth, while the monocular prior restores structure in regions where photometric matching is weak.

\noindent\textbf{MVS-Guided Alignment (MVS-GA).}
The MVS-GA branch aligns the monocular depth to the MVS geometry.
We first apply a global least-squares scale--shift to $\Dmono$ as a robust coarse initialization, and then iteratively refine the remaining local scale mismatch.
To provide strong multi-view geometric evidence, we construct the geometric feature volume
\begin{equation}
\Gmvs = [\CVmulti, \PV, \PVcurve],
\end{equation}
where $\CVmulti$ is the raw multi-view cost volume, $\PV$ is the probability volume, and $\PVcurve$ is a curvature volume derived from $\PV$.
Specifically, $\PVcurve$ is computed from the second-order difference along the depth dimension:
\begin{align}
\Delta_d \PV[d] &= \PV[d{+}1] - \PV[d] - \big(\PV[d] - \PV[d{-}1]\big), \nonumber \\
\PVcurve[d] &= |\Delta_d \PV[d]|.
\end{align}
This curvature term measures the sharpness of the probability peak and thus serves as a confidence proxy:
high curvature corresponds to a clear unimodal response, whereas flat or multi-modal responses often indicate unreliable matching in textureless or occluded regions.

At iteration $k$, the MVS-GA forms the conditioning tensor
$$
\begin{aligned}
\xmvs 
&= [ \Encg(\Gmvs), \Encmvs(\DmvsStep, \CmvsStep), \\
&\quad \Encmono(\DmonoStep), \DmonoStep ],
\end{aligned}
$$
where $\Encg$, $\Encmvs$, and $\Encmono$ are lightweight two-layer convolutional encoders \cite{cheng2025monster}.
Importantly, we concatenate the raw monocular depth $\DmonoStep$ in addition to its encoded features.
This bypass preserves fine-grained geometric detail that may otherwise be oversmoothed by convolutional encoding, allowing the refinement network to directly access precise local depth structure.
For the initial iteration ($k=0$), $\textbf{D}_0^{\text{mvs}}$ and $\textbf{D}_0^{\text{mono}}$ are initialized with the input depth maps, while $\textbf{C}_0^{\text{mvs}}$ is obtained from $\PV$ \cite{cao2024mvsformer++}.

The conditioning feature $\xmvs$ is then processed by a ConvGRU together with the previous hidden state $\hmonoPrev$:
\begin{align*}
\textbf{z}_k &= \sigma( \text{Conv}([\hmonoPrev, \xmvs], \textbf{W}_z) + \textbf{c}_z ), \\
\textbf{r}_k &= \sigma( \text{Conv}([\hmonoPrev, \xmvs], \textbf{W}_r) + \textbf{c}_r ), \\
\hmonoCand &= \tanh( \text{Conv}([\textbf{r}_k \odot \hmonoPrev, \xmvs], \textbf{W}_h) + \textbf{c}_h ), \\
\hmono &= (1 - \textbf{z}_k) \odot \hmonoPrev + \textbf{z}_k \odot \hmonoCand,
\end{align*}
where $\textbf{W}_{*}$ and $\textbf{c}_{*}$ denote learned convolutional kernels and biases, and $\odot$ is element-wise multiplication.
A lightweight prediction head $f_{\theta}$ regresses the residual update
$$
\DeltaDmono = f_{\theta}(\hmono), \quad \textbf{D}_{k+1}^{\text{mono}} = \DmonoStep + \DeltaDmono.
$$
Therefore, MVS-GA does not merely apply a global alignment, but progressively corrects spatially varying monocular scale errors using multi-view geometric cues.

\noindent\textbf{Mono-Guided Refinement (Mono-GR).}
Complementary to MVS-GA, the Mono-GR branch refines the MVS depth using the scale-aligned monocular structure.
This branch is particularly effective in low-texture, reflective, or occluded regions where photometric consistency alone is unreliable.
At iteration $k$, the conditioning tensor is defined as
$$
\begin{aligned}
\xmono 
&= [ \Encmono(\DmonoStep), \Enccross(\DmonoStep, \DmvsStep, \CmvsStep), \\
& \quad \FFPN, \Encmvs(\DmvsStep) ],
\end{aligned}
$$
Here, $\Enccross$ encodes the confidence-weighted discrepancy between the two modalities, explicitly exposing where the monocular and MVS predictions disagree.
Unlike stereo mutual refinement, which typically uses direct pairwise residuals, this discrepancy is gated by the current MVS confidence, so that monocular cues dominate primarily where MVS matching is unreliable.
As a result, Mono-GR focuses on resolving cross-modal conflicts in uncertain regions while avoiding unnecessary changes in already reliable areas.

The hidden state $\hmvs$ is updated by a symmetric ConvGRU, and a two-layer CNN head $f_{\phi}$ predicts the MVS residual:
$$
\DeltaDmvs = f_{\phi}(\hmvs), \quad \textbf{D}_{k+1}^{\text{mvs}} = \DmvsStep + \DeltaDmvs.
$$
In parallel, we predict an updated confidence map from the hidden states of both branches:
$$
\CmvsHat = g_{\psi}([\hmvs, \hmono]).
$$
Rather than recomputing confidence independently at each iteration, we maintain it as a persistent state using an exponential moving average (EMA):
$$
\CmvsNext = (1-\alpha) \cdot \CmvsStep + \alpha \cdot \CmvsHat,
$$
where $\alpha \in (0,1)$ is learnable and initialized to $0.2$ to favor temporal stability in early iterations.
This persistent confidence modeling is another important distinction from prior stereo-style designs, as it explicitly stabilizes iterative refinement under multi-view uncertainty.

Crucially, the updated confidence also serves as an adaptive anchoring signal.
Pixels with high confidence ($\CmvsStep \ge \tau_{\text{conf}}$) are treated as geometric anchors, and large updates to the MVS depth are discouraged at these locations, forcing the monocular prediction to align with reliable MVS geometry.
Conversely, in low-confidence regions, the monocular prior is allowed to play a larger role in hallucinating missing or weakly constrained structure.
This selective rigidity prevents error propagation from uncertain regions while preserving accurate MVS measurements, which is particularly important under sparse-view or weak-texture conditions.

Unlike the stereo methods \cite{cheng2025monster, wen2025foundationstereo} that rely on rectified image pairs, fixed baselines, and simple left-right consistency, MVS must effectively reason across multiple source views. This involves handling diverse baselines, view-dependent occlusions, and complex uncertainty aggregation. To address these unique challenges, we introduce the MVS-specific refinement signals defined above: the geometric feature volume $\Gmvs$, the curvature-based confidence cue $\PVcurve$, the confidence-gated cross-modal discrepancy encoder $\Enccross$, and a persistent EMA-based confidence update with adaptive anchoring. Together, these components ensure that our refinement process remains robust against sparse-view degradation.

After $K$ refinement iterations, the scale-corrected monocular depth $\textbf{D}_K^{\text{mono}}$ and structurally enhanced MVS depth $\textbf{D}_K^{\text{mvs}}$ are forwarded to the next cascade stage.
There, $\textbf{D}_K^{\text{mvs}}$ is used to generate the next-stage depth hypotheses, while $\textbf{D}_K^{\text{mono}}$ is upsampled and reused as the monocular prior.
Through this iterative bidirectional refinement, MDR achieves robust local scale alignment and geometry completion in a manner specifically tailored to the challenges of multi-view stereo.

%%%%%%%%%%%%%%%%%%%%%%%%%%%%%%%%%%%%%%%%%%%%%%%%

\subsection{Loss Function}
We train the cascade MVS framework using three complementary losses applied at every stage $l$. First, we adopt the standard cross-entropy loss $\mathcal{L}_{\mathrm{CE}}^{(l)}$ to measure depth accuracy at stage $l$, i.e., the divergence between the predicted and ground-truth depth distributions \cite{mao2023cross}.

Second, we introduce a mutual refinement loss $\mathcal{L}_{\mathrm{mut}}^{(l)}$. 
%Because each stage outputs $K$ intermediate MVS depth maps $\{ \textbf{D}_k^{\text{mvs},(l)} \}_{k=1}^{K}$, 
%we supervise all of them with a decayed $\ell_1$ loss with $r = 0.9$, so that later refinements are emphasized \cite{lipson2021raft}:
Because each stage outputs $K$ intermediate MVS depth maps $\{ \textbf{D}_k^{\text{mvs},(l)} \}_{k=1}^{K}$, we supervise all of them with a decayed $\ell_1$ loss ($r = 0.9$) to emphasize later refinements \cite{lipson2021raft}:
\begin{equation}
\mathcal{L}_{\mathrm{mut}}^{(l)} = \frac{1}{K} \sum_{k=1}^{K} r^{\, (K - k)} \,
\left\lVert \textbf{D}_k^{\text{mvs},(l)} - \textbf{D}^{\text{gt},(l)} \right\rVert_1.
\end{equation}

\begin{figure*}[t]
    \centering
    \includegraphics[width=0.9\textwidth]{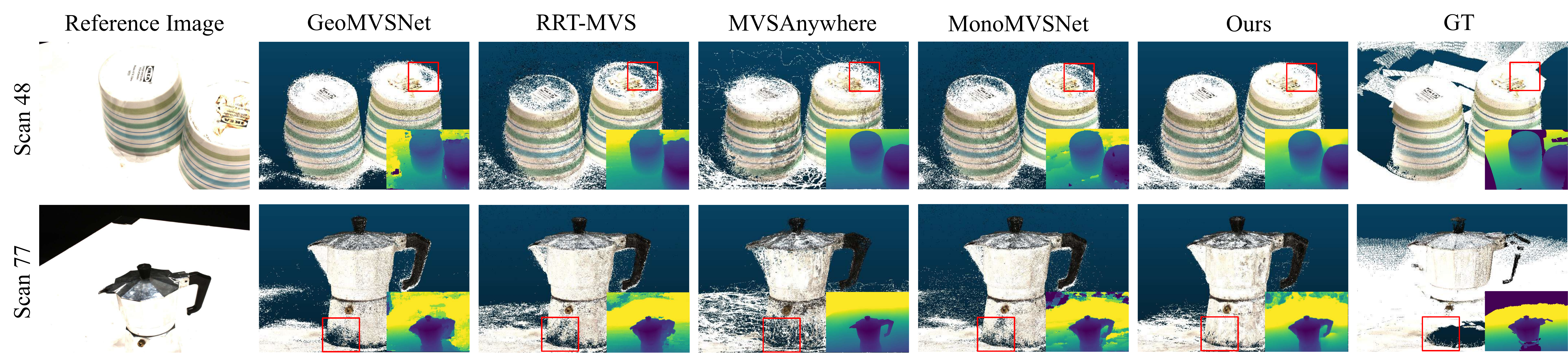}
    \caption{Qualitative results on DTU. Our method produces highly complete 3D reconstructions with clean depth maps even on challenging Scans 48 and 77.}
    \label{fig:fig6}
\end{figure*}

Finally, we add an order-preserving loss $\mathcal{L}_{\text{ord}}^{(l)}$ for the MVS depth map of each stage $l$. Inspired by \cite{lipson2021raft}, to maintain local structure and correct depth ordering, we sample $N_{\text{pair}}$ pixel pairs $(p,q)$ and penalize order violations \cite{chen2016single}:
\begin{equation}
\begin{aligned}
\mathcal{L}_{\text{ord}}^{(l)} 
&= \frac{1}{N_{\text{pair}}}
\sum_{(p,q)}
\max \Big( 0,\,
\big(\textbf{D}^{\text{mvs},(l)}(p) - \textbf{D}^{\text{mvs},(l)}(q)\big) \\
&\hspace{7em}\cdot
\operatorname{sign}\!\big(\textbf{D}^{\text{gt},(l)}(q) - \textbf{D}^{\text{gt},(l)}(p)\big)
\Big),
\end{aligned}
\end{equation}
where $(p,q)$ are randomly sampled pixel pairs from valid depth regions. The $\operatorname{sign}(\cdot)$ function returns $-1$ or $+1$ depending on the relative depth ordering between the two pixels.
The loss is zero when the predicted ordering matches the ground truth and increases linearly otherwise.

The final loss $\mathcal{L}_{\text{total}}$ aggregates all stages:
\begin{equation}
\mathcal{L}_{\text{total}} = 
\sum_{l=1}^{4} 
\big(
\mathcal{L}_{\text{CE}}^{(l)} 
+ \lambda_{\text{mut}} \, \mathcal{L}_{\text{mut}}^{(l)} 
+ \lambda_{\text{ord}} \, \mathcal{L}_{\text{ord}}^{(l)}
\big).
\end{equation}
To balance with the main cross-entropy loss, we set the weights $\lambda_{\text{mut}} = \lambda_{\text{ord}} = 0.02$.

\section{Experiment}
We evaluate the 3D reconstruction performance of our method on the DTU \cite{aanaes2016large} and Tanks and Temples (TNT) \cite{Knapitsch2017} benchmarks, and assess cross-dataset generalization in depth estimation on RobustMVD \cite{schroeppel2022robust}, which provides ground-truth depth for DTU, TNT, and KITTI. Beyond these standard benchmarks, we further examine robustness under sparse-view settings on DTU using only three unfavorable input views. We also conduct ablation studies to validate the contribution of each component and the behavior of our method.

%========================
% Table 1: Normal
%========================
%!FIX (Memory, Time Update, 측정 후 업데이트) -> Ablation이랑 중복되지 않은지 확인
\begin{table}[t]
\centering
\caption{Quantitative results on the DTU evaluation set measured by point-cloud distance metrics (Acc., Comp., and Overall) in millimeters, along with runtime and GPU memory usage.}
\label{tab:results_normal}
\begin{adjustbox}{max width=\columnwidth}
\begingroup
\renewcommand{\arraystretch}{1.05}
\begin{adjustbox}{max width=\textwidth}
\begin{tabular}{l|c|ccccc}
\hline
Methods & Years & Acc.$\downarrow$ & Comp.$\downarrow$ & Overall$\downarrow$ & Time(s)$\downarrow$ & Mem(GiB)$\downarrow$ \\ \hline
CDS-MVSNet~\cite{giang2021curvature}         & ICLR'22 & 0.351 & 0.278 & 0.315 & \cellcolor{yellow!35}0.11 & 2.88 \\
IterMVS~\cite{wang2022itermvs}               & CVPR'22 & 0.373 & 0.354 & 0.363 & \cellcolor{red!35}0.07 & \cellcolor{red!35}0.59 \\
Effi-MVS~\cite{wang2022efficient}            & CVPR'22 & 0.321 & 0.313 & 0.317 & \cellcolor{orange!25}0.08 & \cellcolor{orange!25}1.13 \\
MVSFormer~\cite{cao2022mvsformer}            & TMLR'23 & 0.327 & \cellcolor{yellow!35}0.251 & 0.289 & 0.20 & 1.93 \\
MVSFormer++~\cite{cao2024mvsformer++}        & ICLR'24 & \cellcolor{orange!25}0.309 & 0.252 & \cellcolor{yellow!35}0.281 & 0.21 & 2.65 \\
RRT-MVS~\cite{jiang2025rrt}                  & AAAI'25 & \cellcolor{orange!25}0.309 & 0.261 & 0.285 & 0.32 & \cellcolor{yellow!35}1.80 \\
CF-MVSNet~\cite{ma2024confident}  & TMM'25 & 0.315 & 0.277 & 0.296 & - & -  \\
\hline
Murre~\cite{guo2025multi}                    & CVPR'25 & 1.324 & 0.837 & 1.081 & 5.70 & 8.22 \\
Marigold-DC~\cite{viola2025marigold}         & ICCV'25 & 0.913 & 0.779 & 0.846 & 29.03 & 2.17 \\
MVSAnywhere~\cite{izquierdo2025mvsanywhere}  & CVPR'25 & 0.845 & 0.625 & 0.735 & 0.52  & 4.10  \\
MonoMVSNet~\cite{jiang2025monomvsnet}        & ICCV'25 & \cellcolor{yellow!35}0.313 & \cellcolor{orange!25}0.243 & \cellcolor{orange!25}0.278 & 0.24 & 2.36 \\
Ours                                          & -    & \cellcolor{red!35}0.301 & \cellcolor{red!35}0.241 & \cellcolor{red!35}0.271 &  0.41 &  2.53 \\ \hline
\end{tabular}
\end{adjustbox}
\endgroup
\end{adjustbox}
\end{table}

\subsection{Implementation Details}

\noindent\textbf{Hyperparameters.}
Our model follows a four-stage coarse-to-fine pipeline with inverse depth sampling of 8, 8, 4, and 4 hypotheses from the first to the fourth stage, respectively. The feature dimensionality at each stage is set to 64, 32, 16, and 8. We use $M{=}10$ depth bin masks in Cost-Crafter, and the number of mutual refinement iterations $K$ is set to [12, 8, 5, 3] across the four stages, respectively.

\noindent\textbf{Training.}
For fair comparison, we train and evaluate on the DTU dataset \cite{aanaes2016large} using the standard data splits and view selection protocol \cite{yao2018mvsnet, cao2024mvsformer++}. During DTU training, the model takes 5 input views at a resolution of $512 \times 640$. We adopt the same data augmentation strategy as MVSFormer++ \cite{cao2024mvsformer++}, including random photometric augmentation and spatial transformations.
% Optimization is performed using Adam \cite{kingma2014adam} for 10 epochs with a batch size of 4. % and an initial learning rate of $1\times10^{-3}$, which is decayed by a factor of 0.5 at the 4th, 6th, and 8th epochs.
After DTU pre-training, the model is fine-tuned on the BlendedMVS dataset \cite{yao2020blendedmvs} using 11 input views at a resolution of $576 \times 768$ for 15 epochs with a batch size of 2.
%The learning rate is initialized again to $1\times10^{-3}$ and halved at the 6th, 8th, 10th, and 12th epochs.
Model training is conducted on an RTX A6000 GPU.

\noindent\textbf{Testing.}
We fuse depth maps into 3D point clouds using the dynamic fusion strategy of \cite{yan2020dense}. For DTU evaluation, inference is performed at a resolution of $832 \times 1152$, and the resulting fused point clouds are used for quantitative evaluation. For the TNT benchmark \cite{Knapitsch2017}, following \cite{cao2024mvsformer++}, we perform inference at $1920 \times 1088$ and apply the same fusion pipeline to obtain the final 3D reconstructions. On the RobustMVD benchmark \cite{schroeppel2022robust}, we directly evaluate the pre-trained model without any additional fine-tuning. All testing experiments are conducted on a workstation equipped with an Intel i9-13900KS CPU and an NVIDIA GeForce RTX 3090 Ti GPU. Unless otherwise noted, the runtimes reported in Tables~I and~V measure per-sample depth inference. In contrast, the runtime in Table~IV encompasses the full pipeline, including multi-view depth map fusion and benchmark-specific post-processing.

\begin{table*}[t]
\centering
\caption{Quantitative F-score results on the TNT benchmark. Our method achieves the highest mean F-scores on both test sets.}
\label{tab:quantitative_results}
\resizebox{\textwidth}{!}{%
\begin{tabular}{l|c|c| c c c c c c c c|c| c c c c c c}
\hline
\multirow{2}{*}{\centering\textbf{Methods}}
  & \multirow{2}{*}{\centering\textbf{Years}}
  & \multicolumn{9}{c|}{\rule{0pt}{2.0ex}\textbf{Intermediate}}
  & \multicolumn{7}{c}{\textbf{Advanced}} \\
\cline{3-18}
\rule{0pt}{2ex} 
  & 
  & Mean & Fam. & Fra. & Hor. & Lig. & M60 & Pan. & Pla. & Tra.
  & Mean & Aud. & Bal. & Cou. & Mus. & Pal. & Tem. \\
\hline
\rule{0pt}{2.0ex}COLMAP~\cite{schonberger2016pixelwise} 
  & CVPR'16
  & 42.14 & 50.41 & 22.25 & 25.63 & 56.43 & 44.83 & 46.97 & 48.53 & 42.04 
  & 27.24 & 16.02 & 25.23 & 34.70 & 41.51 & 18.05 & 27.94 \\
CasMVSNet\cite{gu2020cascade} 
  & CVPR'20
  & 56.84 & 76.37 & 58.45 & 46.26 & 55.81 & 56.11 & 54.06 & 58.18 & 49.51 
  & 31.12 & 19.81 & 38.46 & 29.10 & 43.87 & 27.36 & 28.11 \\
%CostFormer\cite{chen2023costformer} 
%  & IJCAI'23
%  & 64.51 & 81.31 & 65.65 & 55.57 & 63.46 & 66.24 & 65.39 & 61.27 & 57.30 
%  & 39.43 & 29.18 & 45.21 & 39.88 & 53.38 & 34.07 & 34.87 \\
% TransMVSNet\cite{ding2022transmvsnet} 
%   & CVPR'22
%   & 63.52 & 80.92 & 65.83 & 56.94 & 62.54 & 63.06 & 60.00 & 60.20 & 58.67 
%   & 37.00 & 24.84 & 44.59 & 34.77 & 46.49 & 34.69 & 36.62 \\
WT-MVSNet\cite{liao2022wt} 
  & NeurIPS'22
  & 65.34 & 81.87 & 67.33 & 57.76 & 64.77 & \cellcolor{yellow!35}65.68 & 64.61 & 62.35 & 58.38 
  & 39.91 & 29.20 & 44.48 & 39.55 & 53.49 & 34.57 & 38.15 \\
RA-MVSNet\cite{zhang2023multi} 
  & CVPR'23
  & 65.72 & 82.44 & 66.61 & 58.40 & 64.78 & \cellcolor{red!35}67.14 & 65.60 & 62.74 & 58.08 
  & 39.93 & 29.14 & 46.04 & 40.30 & 53.22 & 34.63 & 36.28 \\
% DMVSNet\cite{ye2023constraining} 
%   & ICCV'23
%   & 64.66 & 81.27 & 67.54 & 59.10 & 63.12 & 64.64 & 64.80 & 59.83 & 56.97 
%   & 41.17 & 30.08 & 46.10 & 40.65 & 53.53 & 35.08 & 41.60 \\
MVSFormer\cite{cao2022mvsformer} 
  & TMLR'23
  & 66.37 & 82.06 & 69.34 & 60.49 & 68.61 & 65.67 & 64.08 & 61.23 & 59.53 
  & 40.87 & 28.22 & 46.75 & 39.30 & 52.88 & 35.16 & 42.95 \\
MVSFormer++\cite{cao2024mvsformer++} 
  & ICLR'24
  & 67.18 & \cellcolor{red!35}82.69 & 69.44 & \cellcolor{orange!25}64.24 & 69.16 & 64.13 & 66.43 & 61.19 & 60.12 
  & 41.60 & 29.93 & 45.69 & 39.46 & 53.58 & 35.56 & 45.39 \\
RRT-MVS\cite{jiang2025rrt} 
  & AAAI'25
  & \cellcolor{yellow!35}68.16 & \cellcolor{yellow!35}82.54 & \cellcolor{yellow!35}72.31 & 61.44 & \cellcolor{yellow!35}69.89 & 65.32 & \cellcolor{orange!25}68.88 & \cellcolor{yellow!35}64.45 & 60.48 
  & \cellcolor{yellow!35}43.29 & \cellcolor{yellow!35}30.95 & 46.42 & 41.13 & \cellcolor{yellow!35}55.46 & \cellcolor{red!35}37.63 & \cellcolor{orange!25}48.12 \\
GoMVS\cite{wu2024gomvs} 
  & CVPR'24
  & 66.44 & \cellcolor{orange!25}82.68 & 69.23 & \cellcolor{red!35}69.19 & 63.56 & 65.13 & 62.10 & 58.81 & \cellcolor{orange!25}60.80
  & 43.07 & \cellcolor{red!35}35.52 & \cellcolor{orange!25}47.15 & \cellcolor{yellow!35}42.52 & 52.08 & \cellcolor{yellow!35}36.34 & 44.82 \\ 
CF-MVSNet\cite{ma2024confident} 
  & TMM'25
  & 59.78 & 76.89 & 63.52 & 50.62 & 58.75 & 59.10 & 54.31 & 59.70 & 55.31 
  & - & - & - & - & - & - & - \\ 
MonoMVSNet\cite{jiang2025monomvsnet} 
  & ICCV'25
  & \cellcolor{orange!25}68.63 & 82.38 & \cellcolor{red!35}72.89 & 62.80 & \cellcolor{orange!25}70.49 & \cellcolor{orange!25}65.79 & \cellcolor{yellow!35}68.54 & \cellcolor{orange!25}65.54 & \cellcolor{yellow!35}60.59 
  & \cellcolor{orange!25}43.58 & 30.33 & \cellcolor{yellow!35}46.76 & \cellcolor{orange!25}42.90 & \cellcolor{red!35}56.31 & \cellcolor{orange!25}37.28 & \cellcolor{yellow!35}47.88 \\ \hline
Ours 
  & -
  & \cellcolor{red!35}69.01 & \cellcolor{yellow!35}82.54 & \cellcolor{orange!25}72.71 & \cellcolor{yellow!35}63.51 & \cellcolor{red!35}70.63 & 65.30 & \cellcolor{red!35}70.23 & \cellcolor{red!35}65.61 & \cellcolor{red!35}61.54
  & \cellcolor{red!35}43.84 & \cellcolor{orange!25}31.38 & \cellcolor{red!35}47.25 & \cellcolor{red!35}42.81 & \cellcolor{orange!25}55.69 & \cellcolor{red!35}37.63 & \cellcolor{red!35}48.26 \\ \hline
\end{tabular}
}
\end{table*}

\begin{table}[t]
\centering
\caption{Quantitative results on the RobustMVD benchmark, reported using Absolute Relative Error (rel) and Inlier Ratio ($\tau$).}
\label{tab:robustmvd_results}
\setlength{\tabcolsep}{3.2pt}
\renewcommand{\arraystretch}{1.05}
\resizebox{\columnwidth}{!}{%
\begin{tabular}{l c c c c c c c}
\toprule
\multirow{2}{*}[-0.5ex]{Model} & \multirow{2}{*}[-0.5ex]{Years} & \multicolumn{2}{c}{DTU} & \multicolumn{2}{c}{TNT} & \multicolumn{2}{c}{KITTI} \\
\cmidrule(lr){3-4} \cmidrule(lr){5-6} \cmidrule(lr){7-8}
& & rel $\downarrow$ & $\tau \uparrow$ & rel $\downarrow$ & $\tau \uparrow$ & rel $\downarrow$ & $\tau \uparrow$ \\
\midrule
Depth Pro \cite{bochkovskii2024depth} & ICLR'24 & 5.6 & 49.6 & 5.6 & 57.5 & 6.1 & 39.6 \\
DepthAnythingV2 \cite{yang2024depth2} & NeurIPS'24 & 2.6 & 74.7 & 4.5 & 57.5 & 6.6 & 38.6 \\
% PatchmatchNet \cite{wang2021patchmatchnet} & CVPR'21 & 2.1 & 82.8 & 4.8 & 82.9 & 10.8 & 45.8 \\
MVSFormer++ \cite{cao2024mvsformer++} & ICLR'24 & \cellcolor{orange!25}0.9 & \cellcolor{yellow!35}95.3 & 3.2 & 88.1 & 4.4 & 65.7 \\
MonoMVSNet \cite{jiang2025monomvsnet} & ICCV'25 & \cellcolor{orange!25}0.9 & \cellcolor{orange!25}95.5 & \cellcolor{yellow!35}2.6 & \cellcolor{yellow!35}89.0 & \cellcolor{yellow!35}4.1 & \cellcolor{yellow!35}66.2 \\
MVSAnywhere \cite{izquierdo2025mvsanywhere} & CVPR'25 & \cellcolor{yellow!35}1.3 & 95.0 & \cellcolor{orange!25}2.1 & \cellcolor{orange!25}90.5 & \cellcolor{red!35}3.2 & \cellcolor{red!35}68.8 \\
\midrule
Ours & -- & \cellcolor{red!35}0.8 & \cellcolor{red!35}96.2 & \cellcolor{red!35}2.0 & \cellcolor{red!35}90.7 & \cellcolor{orange!25}3.7 & \cellcolor{orange!25}66.8 \\
\bottomrule
\end{tabular}%
}
\end{table}

\subsection{Benchmark Performance}
\noindent \textbf{Evaluation on DTU Dataset.} Table~\ref{tab:results_normal} reports quantitative results on the DTU evaluation set using standard point-cloud metrics (accuracy, completeness, and overall error), alongside runtime and GPU memory usage. Compared to conventional MVS \cite{giang2021curvature, wang2022itermvs, wang2022efficient, cao2022mvsformer, cao2024mvsformer++, jiang2025rrt} and recent DFM-based approaches \cite{guo2025multi, viola2025marigold, izquierdo2025mvsanywhere, jiang2025monomvsnet}, our method achieves the best overall performance, obtaining the lowest errors in accuracy, completeness, and overall. The notable improvement in completeness indicates that our DFM-guided design effectively recovers missing or ambiguous structures, yielding denser reconstructions. Conversely, the DFM-centric method (MVSAnywhere \cite{izquierdo2025mvsanywhere}) and depth-completion approaches (Murre \cite{guo2025multi} and Marigold-DC \cite{viola2025marigold}) exhibit clearly inferior performance. This suggests that relying heavily on foundation models for direct depth prediction is less effective than preserving the explicit geometric constraints of a conventional MVS pipeline.

The MVS-centric baseline MonoMVSNet \cite{jiang2025monomvsnet} achieves strong results, confirming that integrating monocular priors into an explicit multi-view matching framework is superior to purely DFM-driven depth prediction. Nevertheless, our method consistently outperforms MonoMVSNet across all three metrics. This demonstrates our framework leverages monocular priors more effectively, not merely as auxiliary cues, but as structured guidance for both cost-volume construction and mutual depth refinement throughout the cascade MVS process.

Although our method incurs a noticeable runtime increase, it maintains comparable memory usage. More importantly, as discussed in Sect.~IV-C, this added cost is justified by substantial generalization gains over MonoMVSNet. Under constrained sparse-view conditions where baseline performance typically degrades, our method remains robust and better preserves reconstruction quality. This highlights that our bidirectional design generalizes well beyond standard evaluation settings. Fig.~\ref{fig:fig6} shows our method reconstructs finer, more detailed point clouds, particularly in challenging regions.

\noindent\textbf{Evaluation on TNT}. To assess the generalization capability of our method, we evaluate its reconstruction performance using F-scores on the TNT benchmark. Table~\ref{tab:quantitative_results} presents the quantitative results for both the intermediate and advanced sets. Our method achieves the highest mean F-scores on both sets, demonstrating its strong generalization ability.

\noindent\textbf{Evaluation on RobustMVD.} To assess the generalization of our method in depth estimation, we evaluate it on the RobustMVD benchmark. We compare our method against DFMs (Depth Pro \cite{bochkovskii2024depth}, DepthAnythingV2 \cite{yang2024depth2}), an MVS method (MVSFormer++\cite{cao2024mvsformer++}), and the hybrid approaches (MonoMVSNet \cite{jiang2025monomvsnet}, MVSAnywhere \cite{izquierdo2025mvsanywhere}). Following \cite{schroeppel2022robust}, Table~\ref{tab:robustmvd_results} reports two metrics comparing the predicted depth $\hat{d}$ and ground-truth depth $d$: (1) Absolute Relative Error (rel), computed as $|\hat{d}-d|/d$ per pixel, and (2) Inlier Ratio ($\tau$), the percentage of pixels where $\max\!\left(\tfrac{d}{\hat{d}}, \tfrac{\hat{d}}{d}\right) < 1.03$.

Although aligned to the ground-truth scale via post-processing, pure monocular DFM estimates still exhibit the lowest depth estimation performance. %MVS methods perform well on the DTU dataset, where they were trained, but degrade significantly on other datasets. Similarly,
MonoMVSNet relies heavily on MVS; consequently, even when leveraging DFM, its performance remains comparable to pure MVS. In contrast, MVSAnywhere focuses more on DFM, resulting in better depth estimation generalization. Our method achieves the best results on both the DTU and TNT benchmarks. On the KITTI dataset, it performs slightly worse than MVSAnywhere due to challenges with dynamic objects, though the gap is minimal. In summary, while pure MVS and MVS-centric hybrids (MonoMVSNet) excel at 3D reconstruction but falter in depth estimation, and DFM-centric hybrids (MVSAnywhere) show the opposite, our method excels in both tasks, achieving best overall performance.

\begin{figure*}[t]
    \centering
    \includegraphics[width=0.9\textwidth]{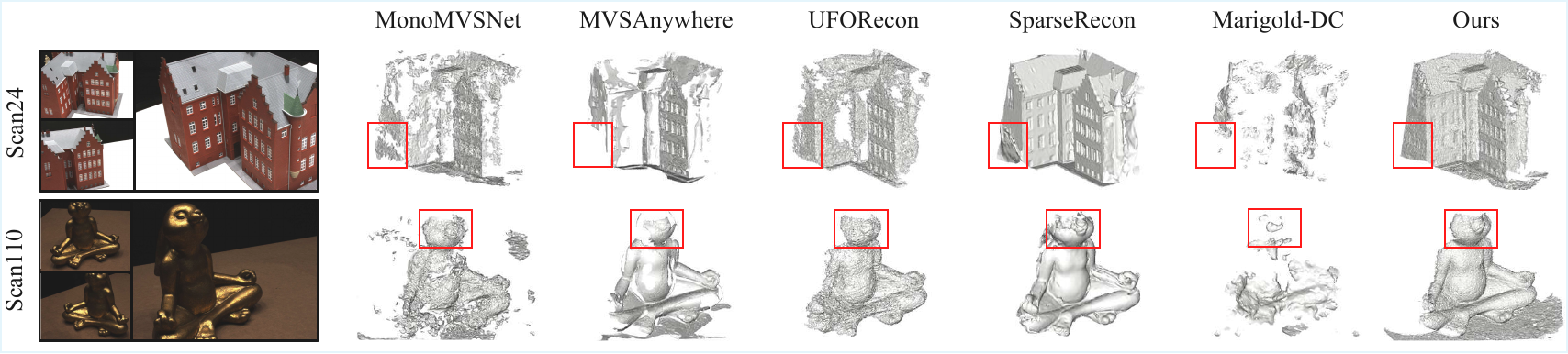}
    \caption{Qualitative sparse-view reconstruction results on DTU under the SparseRecon protocol. Our method recovers more complete and cleaner geometry than competing methods, especially in the highlighted challenging regions.}
    \label{fig:fig7}
\end{figure*}

\begin{table}[t]
\centering
\caption{Sparse-view results under the SparseRecon protocol (3 unfavorable views), evaluated using Chamfer Distance (CD) on the five DTU test scenes. All methods are evaluated at an input resolution of $576\times768$, and runtime (s) and GPU memory (GiB) are reported under the same setting.}
\resizebox{\columnwidth}{!}{%
\begin{tabular}{l l| c| c c c c c c| c c}
\toprule
\multicolumn{2}{c|}{\multirow{2}{*}{Methods}} & 
\multirow{2}{*}{Years} & 
\multicolumn{6}{c|}{Chamfer Distance $\downarrow$} & 
\multirow{2}{*}{Time $\downarrow$} & 
\multirow{2}{*}{Mem $\downarrow$} \\
\cline{4-9}
& & & \rule{0pt}{2.2ex}24 & 34 & 110 & 114 & 118 & Avg. $\downarrow$ & & \\  \hline
% \multicolumn{10}{c}{General MVS} \\ \hline
\multicolumn{2}{l|}{{\scriptsize{\color{gray}{General MVS:}}}} & & & & & & & & &  \\
& CDS-MVSNet~\cite{giang2021curvature} & ICLR'22 & 1.98 & 1.43 & 1.02 & 0.86 & 1.87 & 1.43 & \cellcolor{yellow!35}0.64 & 1.30 \\ 
& IterMVS~\cite{wang2022itermvs} & CVPR'22 & 2.65 & 1.36 & 1.01 & 0.66 & 1.58 & 1.45 & \cellcolor{red!35}0.60 & \cellcolor{red!35}0.27 \\ 
& Effi-MVS~\cite{wang2022efficient}   & CVPR'22 & 4.35 & 1.88 & 3.54 & 1.67 & 3.15 & 2.92 & \cellcolor{orange!25}0.62 & \cellcolor{orange!25}0.51 \\ 
& MVSFormer~\cite{cao2022mvsformer}  & TMLR'23 & \cellcolor{yellow!35}1.34 & \cellcolor{yellow!35}0.74 & \cellcolor{red!35}0.66 & \cellcolor{yellow!35}0.61 & 1.47 & \cellcolor{yellow!35}0.97 & 0.71 & 0.83 \\ 
& MVSFormer++~\cite{cao2024mvsformer++}  & ICLR'24 & 1.51 & 0.81 & \cellcolor{orange!25}0.71 & 0.63 & \cellcolor{yellow!35}1.38 & 1.01 & 0.72 & 1.90 \\ 
& RRT-MVS~\cite{jiang2025rrt}  & AAAI'25 & 3.97 & 1.84 & 2.22 & 1.42 & 2.69 & 2.43 & 0.88 & \cellcolor{yellow!35}0.78 \\ \hline
% \multicolumn{10}{c}{Sparse-View MVS} \\ \hline
\multicolumn{2}{l|}{{\scriptsize{\color{gray}{Sparse-View MVS:}}}} & & & & & & & & &  \\
& UFORecon~\cite{na2024uforecon} & CVPR'24  & 1.52 & 0.79 & 0.93 & 0.66 & \cellcolor{orange!25}1.26 & 1.03 & 140.02 & 3.29 \\ 
& SparseRecon~\cite{han2025sparserecon} & ICCV'25   & \cellcolor{orange!25}1.26 & \cellcolor{orange!25}0.72 & 0.77 & \cellcolor{red!35}0.44 & \cellcolor{red!35}0.83 & \cellcolor{orange!25}0.80 & 11970.40 & 7.04 \\ \hline
% \multicolumn{10}{c}{Depth Completion Model} \\ \hline
% & {\scriptsize{\color{gray}{Depth-Completion:}}} & & & & & & & & &  \\
\multicolumn{2}{l|}{{\scriptsize{\color{gray}{Depth-Completion:}}}} & & & & & & & & &  \\
& Murre~\cite{guo2025multi}  & CVPR'25 &  5.10 & 2.19 & 3.03 &  1.71 & 2.69 & 2.95 & 17.63 & 8.04 \\
& Marigold-DC~\cite{viola2025marigold} & ICCV'25 & 2.90 & 2.24 & 3.44 & 1.67  & 2.90 & 2.63 & 87.60 & 1.99 \\ \hline
% \multicolumn{10}{c}{MVS Using Depth Foundation Model} \\ \hline
% & {\scriptsize{\color{gray}{MVS Using DFM:}}} & & & & & & & & &  \\
\multicolumn{2}{l|}{{\scriptsize{\color{gray}{MVS Using DFM:}}}} & & & & & & & & &  \\
& MVSAnywhere~\cite{izquierdo2025mvsanywhere} & CVPR'25 & 2.34 & 1.53 & 2.47 & 0.78 & 1.67 & 1.76 & 1.09 & 3.75 \\
& MonoMVSNet~\cite{jiang2025monomvsnet}   & ICCV'25 & 2.80 & 1.40 & 1.20 & 0.90 & 1.80 & 1.62 & 0.76 & 1.18 \\
& Ours        &  -   & \cellcolor{red!35}1.22 & \cellcolor{red!35}0.71 & \cellcolor{yellow!35}0.75 & \cellcolor{orange!25}0.46 & \cellcolor{red!35}0.83 & \cellcolor{red!35}0.79 & 0.89 & 1.21 \\
\bottomrule
\end{tabular}%
}
\label{tab:dtu_cd}
\end{table}

\begin{figure}[t]
\centering
\includegraphics[width=0.95\linewidth]{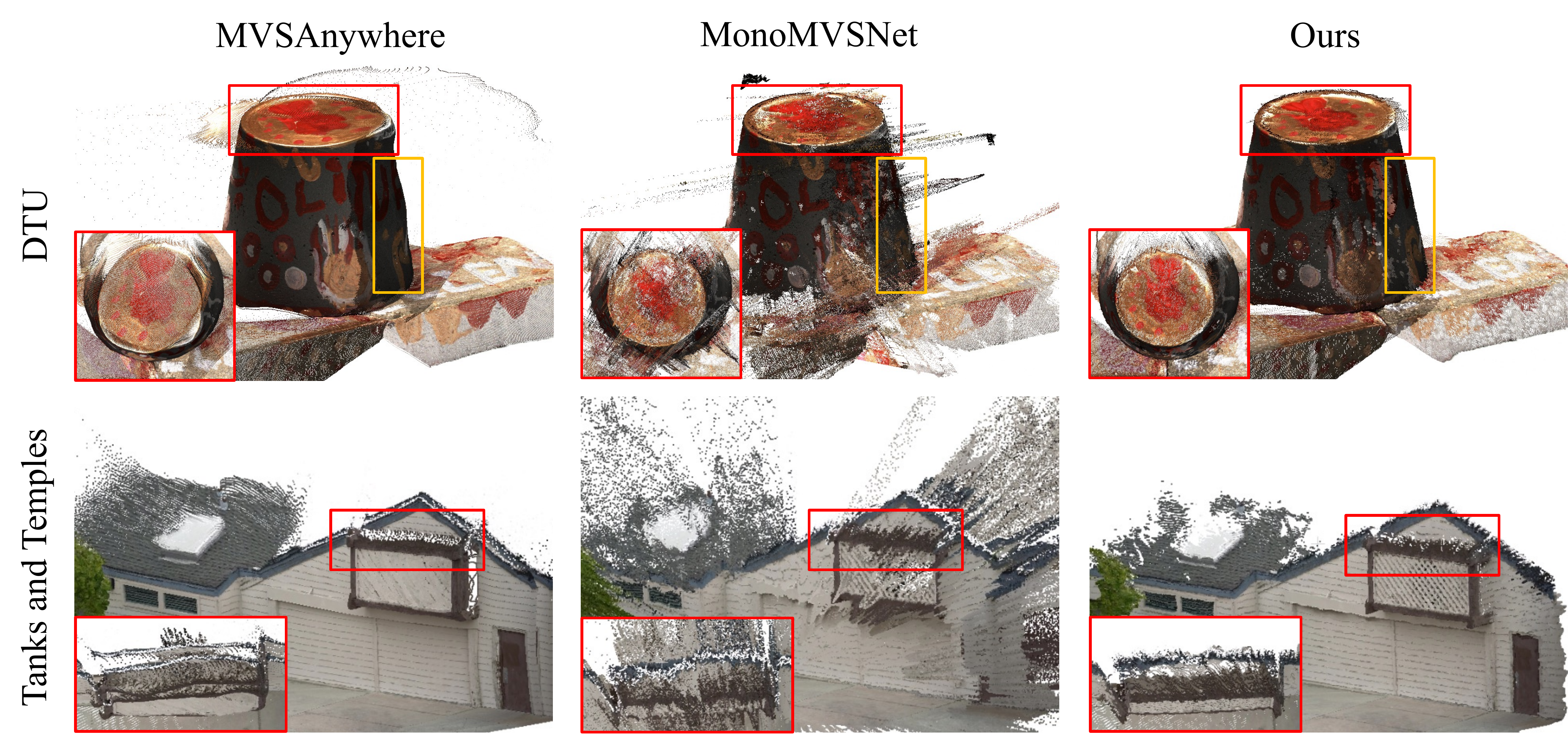}
\caption{Qualitative comparison of two-view reconstruction. Compared to MonoMVSNet, which produces outliers, and MVSAnywhere, which suffers from scale ambiguity, our method generates complete and aligned surfaces.}
\label{fig:fig8}
\end{figure}

\subsection{Generalization to Sparse-View Reconstruction}
\label{sec:sparse_view}

A key advantage of our method is its strong generalization beyond standard MVS to extremely sparse-view reconstruction. To validate this property, we evaluate our method under sparse-view conditions on DTU. We compare it with conventional MVS approaches and sparse-view-targeted methods such as UFORecon \cite{na2024uforecon} and SparseRecon \cite{han2025sparserecon}, which primarily optimize surfaces in a volumetric form similar to SDF.

Unlike these specialized baselines, our approach follows a conventional depth estimation and fusion pipeline, without being specifically designed for sparse-view reconstruction. For a controlled comparison, we adopt the SparseRecon \cite{han2025sparserecon} protocol, using three unfavorable views (22, 25, 28) with limited overlap. However, this serves as a protocol-level evaluation rather than a matched training-setting comparison, as our method and the baselines (UFORecon, SparseRecon) rely on different training splits. Therefore, we restrict evaluation to five DTU test scenes shared across settings.

Table~\ref{tab:dtu_cd} reports sparse-view reconstruction results using Chamfer Distance (CD), where lower values mean higher accuracy. Although not explicitly trained for sparse-view reconstruction, our model achieves the best overall performance with the lowest average CD. It outperforms general MVS methods. Notably, it also achieves lower errors than sparse-view-targeted approaches such as UFORecon and SparseRecon. This demonstrates that our depth-foundation-guided cost construction and mutual depth refinement remain effective even given only three unfavorable views.

While the margin over SparseRecon is modest, our method provides a vastly superior accuracy-efficiency trade-off. Dedicated techniques like UFORecon and SparseRecon rely on intensive volume-rendering-based surface optimization, causing substantial runtime and memory costs. In contrast, our approach achieves better accuracy with minimal computational overhead. Compared to fast MVS baselines, it introduces marginal additional cost while delivering stronger sparse-view capabilities. Qualitative comparisons in Fig.~\ref{fig:fig7} support these quantitative trends. In highlighted regions, our reconstructions preserve object structures more faithfully, exhibiting fewer missing parts and geometric distortions than competitors.
% This confirms our method generalizes robustly to limited-view settings without expensive test-time optimization. 

We also assessed MVS performance using two views. Figs.~\ref{fig:overview} and \ref{fig:fig8} illustrate two-view reconstructions from DFM-based MVS approaches. MonoMVSNet produced significant noise and artifacts in non-overlapping regions. MVSAnywhere generated cleaner models but suffered from scale ambiguity, causing misaligned surfaces. In contrast, our method successfully reconstructed regions without view overlap, producing well-aligned, scale-corrected 3D surfaces. This demonstrates our method effectively integrates DFM and MVS, achieving robust sparse-view reconstructions.

% \newpage ... \newpage

%!FIX (Memory, Time Update, 측정 후 업데이트)
\begin{table}[t]
\centering
\caption{ Ablation study on the components of our method on the DTU dataset: Cost-Crafter (CC), Monocular Volume Fusion (MV), Bin Mask-based Refinement (BM), Mutual Depth Refinement (MDR), Mutual Refinement Loss (Mut), and Order-preserving Loss (Ord). We also report Mean Absolute Error (MAE) in $mm$, and average GPU memory (GiB) and runtime (seconds).}
\label{tab:ablation_study}
\begin{adjustbox}{max width=\columnwidth}
% \setlength{\tabcolsep}{2pt}      % column 사이 간격 줄이기 (기본 6pt → 2pt 정도)
% \renewcommand{\arraystretch}{1.1}% 세로 간격 약간만 늘려서 가독성 확보
% \resizebox{\columnwidth}{!}{%
\begin{tabular}{cccccccccccc} 
\toprule
\multirow{2}{*}[-0.5ex]{Model} & \multicolumn{2}{c}{CC} & \multirow{2}{*}[-0.5ex]{MDR} & \multicolumn{2}{c}{Loss} & \multirow{2}{*}[-0.5ex]{Overall$\downarrow$} & \multirow{2}{*}[-0.5ex]{Acc.$\downarrow$} & \multirow{2}{*}[-0.5ex]{Comp.$\downarrow$} & \multirow{2}{*}[-0.5ex]{MAE$\downarrow$} & \multirow{2}{*}[-0.5ex]{Mem$\downarrow$} & \multirow{2}{*}[-0.5ex]{Time$\downarrow$}\\
\cmidrule(lr){2-3} \cmidrule(lr){5-6} 
 & MV & BM & & Mut & Ord & & & & & &\\ \midrule
A & & & & & & 0.292 & 0.320 & 0.264 & 6.01 & 1.71 & 0.18\\
B & \checkmark & & & & & 0.286 & 0.312 & 0.260 & 5.82 & 2.04 & 0.22\\
C & & \checkmark & & & & 0.287 & 0.315 & 0.259 & 5.86 & 1.99 & 0.21\\
D & \checkmark & \checkmark & & & & 0.283 & 0.309 & 0.256 & 5.64 & 2.23 & 0.23\\
E & & & \checkmark & & & 0.281 & 0.307 & 0.255 & 5.14 & 2.24 & 0.36\\
F & \checkmark & \checkmark & \checkmark & & & 0.275 & 0.303 & 0.247 & 4.91 & 2.53 & 0.41\\
G & \checkmark & \checkmark & \checkmark & \checkmark & & 0.272 & 0.298 & 0.246 & 4.88 & 2.53 & 0.41\\
H & \checkmark & \checkmark & \checkmark & \checkmark & \checkmark & 0.271 & 0.301 & 0.241 & 4.84 & 2.53 & 0.41\\
\bottomrule
\end{tabular}%
\end{adjustbox}
% }
\end{table}

\subsection{Ablation Study}

Table~\ref{tab:ablation_study} details our ablation study on the DTU dataset. Progressively integrating Cost-Crafter, MDR, and auxiliary losses into the baseline (Model-A) consistently improves reconstruction quality. Within Cost-Crafter, monocular volume fusion (Model-B) and bin-mask refinement (Model-C) independently reduce errors. Adding mutual refinement (Model-G) and order-preserving (Model-H) losses achieves optimal accuracy and completeness, maintaining a practical accuracy-efficiency trade-off despite minor overhead.

Comparing Model-D (Cost-Crafter only) and Model-E (MDR only) clarifies their distinct roles. Cost-Crafter ensures strong coarse geometry by injecting monocular priors, but without MDR, it fails to propagate these priors to later stages. This causes noisy results and lower completeness in textureless regions. Conversely, MDR yields smoother, denser depth maps by suppressing artifacts in ambiguous areas. However, lacking Cost-Crafter's reliable initial hypotheses, MDR struggles to rectify early geometric errors, limiting overall precision.

\section{Conclusion}
We introduce a framework tightly unifying DFM monocular depth priors with a cascade MVS pipeline. Unlike approaches using static fusion, ours enables continuous interaction between monocular and multi-view cues via: (i) prior-guided cost refinement, injecting structural priors into initial matching for a stronger geometric foundation, and (ii) bidirectional mutual refinement, iteratively aligning monocular depth with multi-view geometry. Consequently, our framework mitigates monocular scale ambiguity and compensates for MVS incompleteness in challenging areas with occlusions or limited view overlap.

Experiments on DTU, TNT, and RobustMVD show our method consistently surpasses state-of-the-art approaches in 3D reconstruction and depth estimation. Furthermore, it generalizes robustly to sparse-view settings without specific training, competing with dedicated methods while maintaining a superior accuracy-efficiency trade-off. Overall, our approach resolves limitations in prior hybrid methods \cite{jiang2025monomvsnet, izquierdo2025mvsanywhere}, successfully combining monocular structural completeness with MVS metric reliability. This work offers a practical direction for generalized 3D reconstruction leveraging foundation priors without sacrificing geometric consistency.

\clearpage
\setcounter{page}{1}
\noindent{\huge\bfseries Supplementary Material}\par\vspace{0.8em}

\section{Detailed Method}

\subsection{Feature Extraction and Fusion}

Our feature extraction follows the cascade MVS framework of MVSFormer++~\cite{cao2024mvsformer++}, with key modifications for efficiency and integration with monocular priors. We adopt Depth Anything V2-Small~\cite{yang2024depth2} as the DFM encoder, which uses DINOv2's ViT-Small backbone containing approximately $4\times$ fewer parameters than MVSFormer++'s ViT-Base, significantly improving computational efficiency.
Unlike MVSFormer++, which extracts features from both reference and source views, we extract DINOv2 features only from the reference image $\mathbf{I}_0$ to obtain $\mathbf{F}_\text{DINO}$. These reference features are then combined with multi-view FPN features inside the Side View Attention (SVA) module, which injects multi-view geometric consistency into the refined representation.

For multi-scale representation, we employ a Feature Pyramid Network (FPN) to extract features $\{\mathbf{F}^l_{\text{FPN}}\}_{l=1}^4$ at four resolutions. A key distinction from MVSFormer++ is our integration of Frustoconical Positional Encoding (FPE), which encodes 3D spatial information by applying sinusoidal positional encoding along $x$, $y$, and depth dimensions based on the camera frustum geometry. Following MonoMVSNet~\cite{jiang2025monomvsnet}, we fuse FPE directly with the features before cost volume construction, rather than during cost volume regularization. This early geometric encoding ensures that both semantic DINOv2 features and multi-scale FPN features are spatially aligned with 3D scene geometry from the outset.

\subsection{Monocular Volume Construction}

This section details the construction of the monocular cost volume, $\mathbf{CV}_{\text{mono}}$, which serves as a key input to the Cost-Crafter module. Unlike the conventional multi-view cost volume $\mathbf{CV}_{\text{multi}}$, which relies on photometric consistency across views, the monocular volume is derived solely from a single-image depth prior. This representation provides critical structural guidance, particularly in regions where multi-view correspondence is ambiguous.

\begin{figure}[t]
    \centering
    \includegraphics[width=\linewidth]{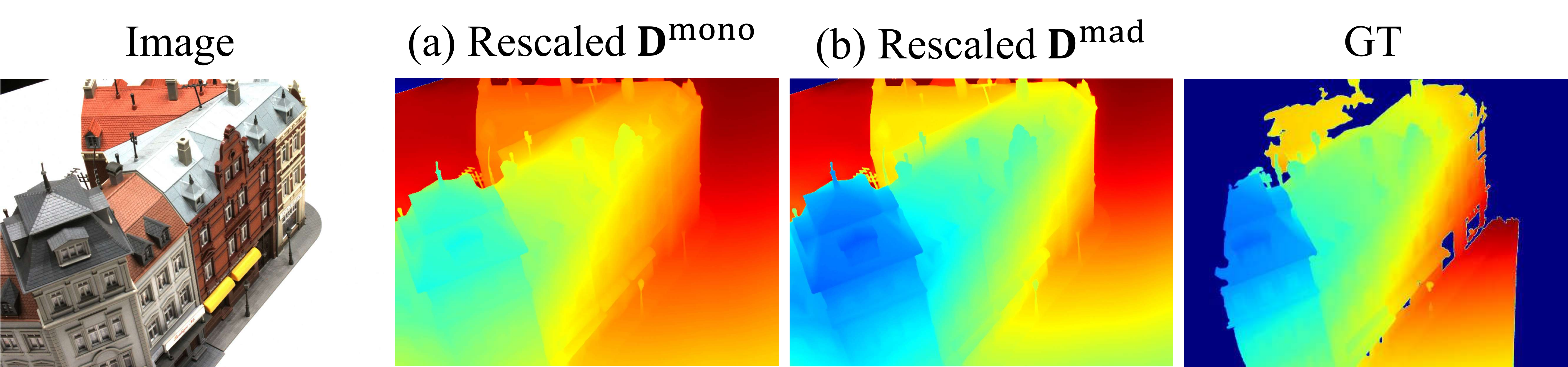}
    \caption{
    Illustration of scale alignment of depth maps. (a) The monocular depth map $\mathbf{D}^{\text{mono}}$ is rescaled to fit within a predefined depth hypothesis range. However, due to the presence of a few outliers, the resulting scale often fails to align accurately with the ground truth. (b) The Median Absolute Deviation (MAD) method removes statistical outliers, resulting in a refined depth map $\mathbf{D}^{\text{mad}}$.
    % By normalizing and rescaling, the values in $\mathbf{D}^{\text{mad}}$, more accurate and consistent scale alignment with the ground-truth depth can be achieved.
    This normalization and rescaling of $\mathbf{D}^{\text{mad}}$ leads to more accurate and consistent scale alignment with the ground-truth depth.
    }
    \label{fig:MAD}
\end{figure}

\noindent\textbf{Depth Refinement and Normalization.}
Given a reference image $\mathbf{I}_{0}$, we first obtain an initial monocular depth map $\mathbf{D}^{\text{mono}}$ using a DFM. To leverage the robust generalization capabilities learned from large-scale diverse datasets, we keep the DFM frozen during training rather than fine-tuning it on our limited domain-specific data.
To mitigate noise inherent in the raw prediction, we apply Median Absolute Deviation (MAD) filtering~\cite{hampel1986robust}. Specifically, we employ an asymmetric variant of MAD that computes separate deviation bounds for the left and right intervals. We empirically found that this asymmetric approach effectively suppresses depth outliers while preserving essential depth discontinuities and fine structural details.
Following filtering, the smoothed depth map $\mathbf{D}^{\text{mad}}$ is normalized to the range $[0, 1]$ and linearly rescaled to align with the scene-specific depth range $[d_{\min}, d_{\max}]$, resulting in the refined map $\mathbf{D}^{\text{mono}}$. As illustrated in Fig.~\ref{fig:MAD}, this refinement process yields a monocular prior that is more accurately aligned with the ground-truth global scale while retaining sharp geometric boundaries.

\noindent\textbf{Gaussian-Weighted Soft Encoding.}
To integrate the continuous monocular depth into the MVS pipeline, we construct $\mathbf{CV}_{\text{mono}}$ by aligning the depth values with the discrete multi-view depth hypotheses $\{d_1, \ldots, d_D\}$, sampled uniformly in the inverse depth space. Instead of using hard one-hot encoding, which rigidly assigns a pixel to a single depth plane, we adopt a Gaussian-weighted soft one-hot representation~\cite{jang2016categorical}.
This soft encoding strategy offers two key advantages: it accounts for local scale inconsistencies where monocular predictions typically deviate slightly from metric truth, and it facilitates smooth gradient propagation for end-to-end optimization.
Formally, for a pixel at spatial location $(h, w)$ with a refined depth value $d$, the probability mass is distributed across the depth hypotheses $d_i$ as follows:
\begin{equation}
\mathbf{CV}_{\text{mono}}[i, h, w] = \frac{1}{Z} \exp\left(-\frac{(d - d_i)^2}{2\sigma^2}\right),
\end{equation}
where $d_i$ denotes the $i$-th depth hypothesis, and $\sigma$ represents the expected local scale uncertainty, set to half the average interval between adjacent hypotheses. The term $Z = \sum_{j=1}^{D} \exp\left(-\frac{(d - d_j)^2}{2\sigma^2}\right)$ serves as a normalization factor. This formulation naturally concentrates the probability density near the predicted depth while allowing a smooth decay for distant hypotheses, thereby modeling the inherent uncertainty in the monocular prior. The resulting volume $\mathbf{CV}_{\text{mono}} \in \mathbb{R}^{1\times D\times{{H_1}}\times{{W_1}}}$ is subsequently fused with $\mathbf{CV}_{\text{multi}}$ via bidirectional cross-attention in the Cost-Crafter module.

\section{More Experiments}

\subsection{Additional Ablation Study}

\begin{figure}[t]
    \centering
    \includegraphics[width=\linewidth]{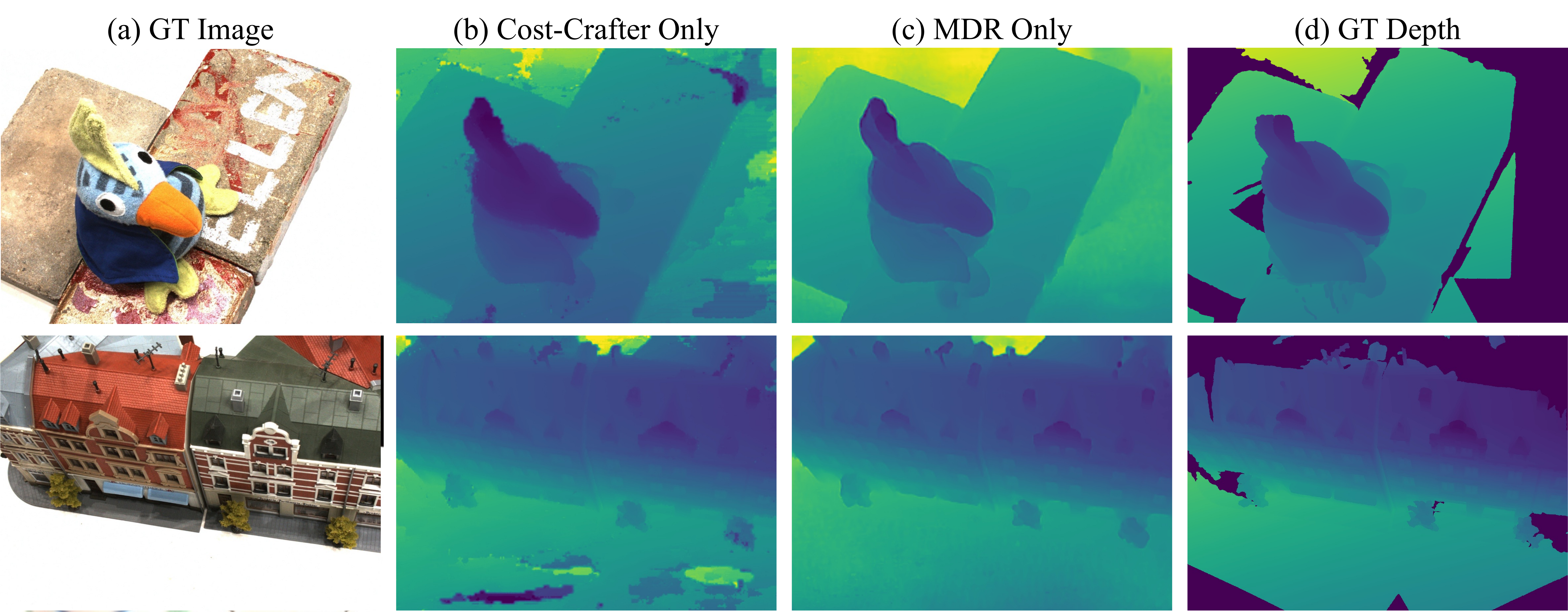}
    \caption{Qualitative visualization of component contributions. The ``Cost-Crafter only'' configuration provides accurate geometric guidance but tends to produce noise in textureless regions. In contrast, the ``MDR only'' configuration generates smoother and more complete depth maps but suffers from scale inaccuracies due to the lack of precise initialization.
    }
    \label{each_model_effect}
\end{figure}

\noindent \textbf{Cost-Crafter vs MDR.} Fig.~\ref{each_model_effect} shows examples of the depth map results for ``Cost-Crafter only'' and ``MDR only'', respectively. As illustrated in Fig.~\ref{each_model_effect}, Cost-Crafter primarily provides accurate geometric initialization, while MDR ensures structural smoothness and completeness. Combining them in the full model leverages these complementary strengths, yielding the most balanced results with both high geometric accuracy and strong reconstruction completeness.

% \begin{table}[t]
% \caption{Quantitative ablation study on the DTU evaluation dataset. We compare the performance of utilizing the Cost-Crafter module independently versus the Mutual Depth Refinement (MDR) module independently. While ``Cost-Crafter only'' excels in accuracy and ``MDR only'' improves completeness, the full model achieves the lowest error rates across all metrics.}
% \resizebox{\columnwidth}{!}{%
% \begin{tabular}{l|cccccc}
% \toprule
% Method & Overall$\downarrow$ & Acc.$\downarrow$  & Comp.$\downarrow$ & 2mm$\downarrow$ & 4mm$\downarrow$ & 8mm$\downarrow$ \\ \midrule
% Baseline  & 0.292   & 0.320 & 0.264 &  15.89 & 10.78 & 7.04 \\
% Cost-Crafter only & 0.285   & 0.311 & 0.257 &  14.06 & 9.75 & 6.30 \\ 
% MDR only         & 0.283   & 0.317 & 0.249 &  14.76 & 10.39 & 5.71 \\
% Full Model  & 0.274   & 0.301 & 0.248 &  13.82 & 8.62 & 5.18 \\ \bottomrule
% \end{tabular}
% }
% \label{table:each_model_effect}
% \end{table}
%%%%%%%%%%%%%%%%%%%%%%%%%%%%%%%%%%%%%%%%%%%%%%%%%%%%%%%%%%%%%%%%%%%%%%%%%%%%%%%%%%%%%%%

\begin{figure}[t]
    \centering
    \includegraphics[width=\linewidth]{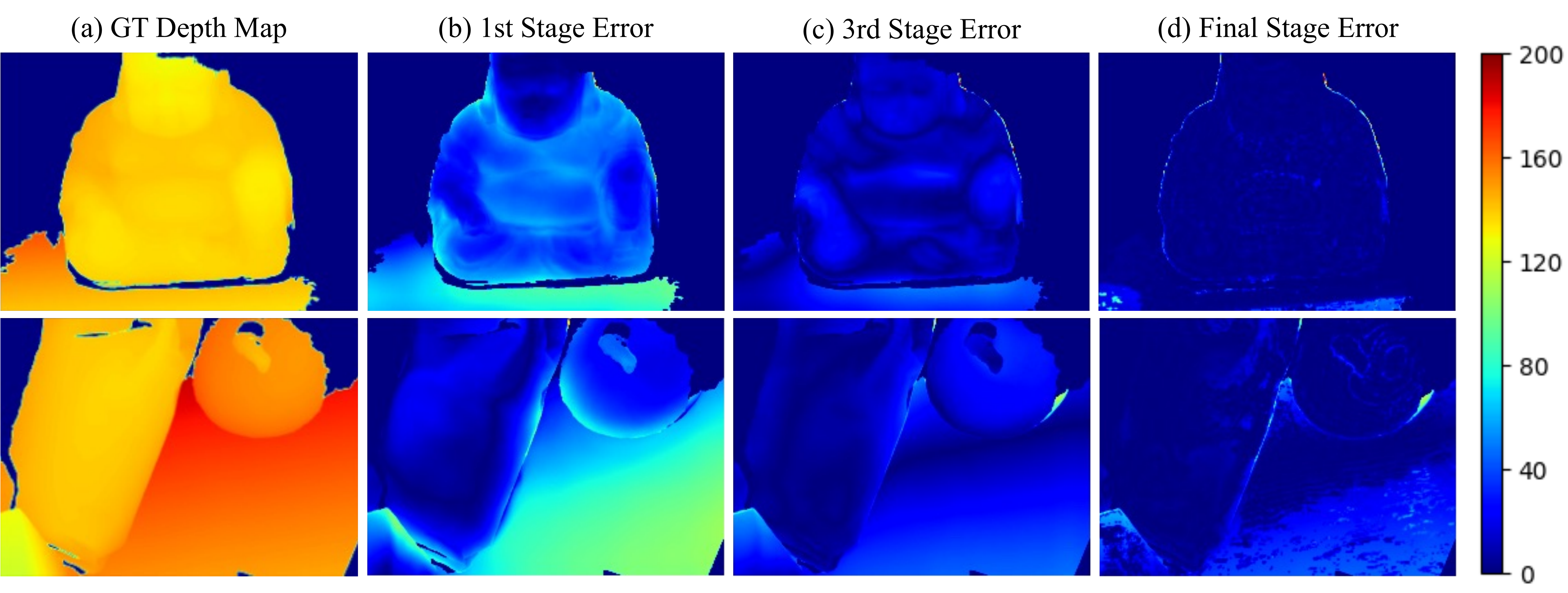}
    \caption{
    Visualization of the progressive refinement of the monocular depth map ($\text{\textbf{D}}^{\text{mono}}$) across cascade stages. (a) Ground Truth depth map. (b)--(d) Error maps representing the absolute difference between the estimated monocular depth and the ground truth at the 1st, 3rd, and Final stages, respectively. As the refinement proceeds through the cascade, the error magnitude significantly decreases, demonstrating that our method iteratively corrects scale ambiguity and enhances geometric consistency to match the ground truth.
    }
    \label{mono_depth_refinement_process}
\end{figure}

\noindent \textbf{Robustness and Progressive Refinement Analysis.}
This section visually evaluates how our MDR module progressively enhances depth accuracy and robustly handles erroneous monocular inputs. Fig.~\ref{mono_depth_refinement_process} validates our iterative scale correction by displaying error maps of the monocular prior ($\text{\textbf{D}}^{\text{mono}}$) at the 1st, 3rd, and Final stages. Initially, high error magnitudes appear due to inherent scale ambiguity. However, as refinement proceeds, errors decrease significantly, converging toward the ground truth. The final map confirms that our coarse-to-fine, bidirectional strategy enforces geometric consistency, aligning the monocular prior with the true metric scale.

\begin{figure}[t]
    \centering
    \includegraphics[width=\linewidth]{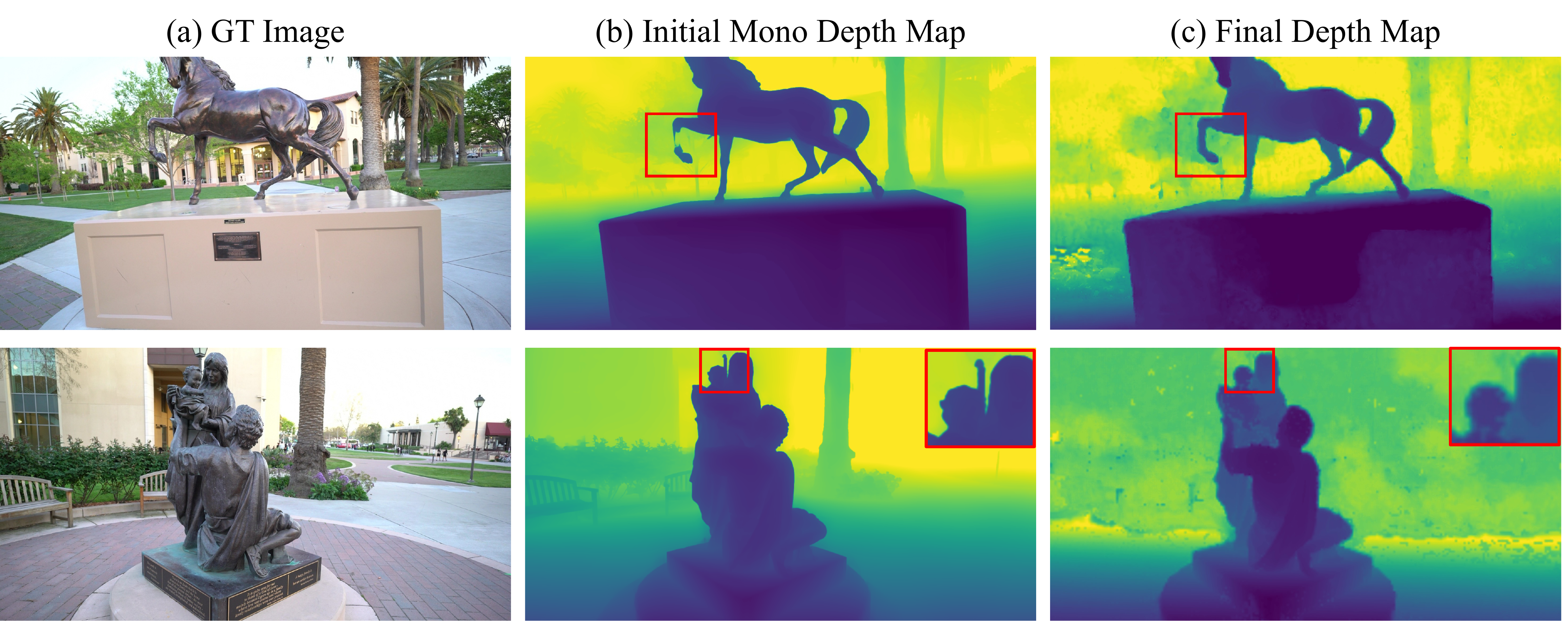}
    \caption{Visual demonstration of robustness against erroneous DFM priors. (a) GT Image. (b) Initial monocular depth map estimated by the DFM, showing structural artifacts (highlighted in red boxes). (c) Final depth map output. Even when the initial monocular prior contains significant errors, our method leverages multi-view geometric consistency to effectively identify and override these inaccuracies, resulting in a geometrically correct final depth estimation.}
    \label{wrong_dfm}
\end{figure}

Beyond scale correction, MDR prevents error propagation from structurally flawed DFM predictions by leveraging multi-view consistency. As shown in Fig.~\ref{wrong_dfm}, initial DFM predictions often exhibit structural errors due to semantic ambiguity, such as omitting a statue's leg or merging figures. Despite these inaccuracies, our method successfully recovers the correct geometry. This demonstrates that our bidirectional interaction avoids blindly trusting the prior; instead, high-confidence MVS geometric cues override the DFM's structural flaws.

\noindent \textbf{Robustness to the Number of Source Views.}
Table~\ref{tab:number_of_sources} compares our method with the baseline~\cite{cao2024mvsformer++} across varying source views ($N \in \{3, 5, 7, 10\}$), reporting 3D reconstruction quality, depth error rates ($>2$\,mm and $>4$\,mm), and computational cost on DTU. Decreasing $N$ from 5 to 3 severely impacts the baseline, increasing its overall and $>2$\,mm depth errors by 4.1\% and 19.4\%, respectively. Conversely, our method demonstrates stronger sparse-view robustness, limiting these increases to 3.6\% and 11.3\%. This confirms the monocular prior effectively compensates for reduced multi-view overlap, preserving depth stability. Furthermore, while increasing $N$ to 10 reduces the baseline's overall error by 2.6\%, our method saturates early at $N=5$, yielding a negligible 0.1\% improvement. This demonstrates our approach achieves near-optimal performance without relying on dense view configurations.

\begin{table*}[ht]
\centering
\caption{
Quantitative results for the baseline and our method with varying numbers of source views $N$ on the DTU test set. 3D reconstruction errors are reported using Accuracy, Completeness, and Overall scores (lower is better). Depth errors are measured by the percentage of pixels deviating more than 2mm ($e_2$) or 4mm ($e_4$). Values in parentheses indicate the relative performance change compared to the default setting ($N=5$). GPU memory usage and inference time are also reported.
}
\setlength{\tabcolsep}{1mm}{
\resizebox{\textwidth}{!}{%
\begin{tabular}{c| c| l l l| l l| c| c}
\hline
% \rule{0pt}{2.0ex}
Method & Num. Views & Acc.$\downarrow$ & Comp.$\downarrow$ & Overall$\downarrow$ & $e_2\downarrow$ & $e_4\downarrow$ & GPU[GiB]$\downarrow$ & Time(s)$\downarrow$\\
\hline
\rule{0pt}{2.0ex}

\multirow{4}{*}{Baseline}
& 3   & 0.3308 (2.5\%$\uparrow$)      & 0.2788 (6.1\%$\uparrow$)     & 0.3048 (4.1\%$\uparrow$)      & 
19.01 (19.4\%$\uparrow$)  & 12.13 (21.4\%$\uparrow$) &   2.72 & 0.16 \\ 
& \textbf{5} & \textbf{0.3227} & \textbf{0.2628} & \textbf{0.2928} & \textbf{15.92} & \textbf{9.99} & \textbf{2.88}  & \textbf{0.24} \\
& 7   & 0.3201 (0.8\%$\downarrow$) & 0.2612 (0.6\%$\downarrow$) & 0.2909 (0.7\%$\downarrow$)& 15.20 (4.7\%$\downarrow$)  &  9.51 (5.0\%$\downarrow$) &   3.00 & 0.31 \\
& 10  & 0.3033 (6.4\%$\downarrow$) & 0.2583  (1.7\%$\downarrow$) & 0.2852 (2.6\%$\downarrow$)& 13.63 (16.8\%$\downarrow$)  &  8.73 (14.4\%$\downarrow$) &  3.19  & 0.42 \\
\hline
% \rule{0pt}{2.0ex}
\multirow{4}{*}{Ours}
& 3          & 0.3080 (2.2\%$\uparrow$) & 0.2544 (5.3\%$\uparrow$) & 0.2812 (3.6\%$\uparrow$) & 15.38 (11.3\%$\uparrow$)& 9.89 (14.7\%$\uparrow$) & 2.42 & 0.34 \\
& \textbf{5} & \textbf{0.3013} & \textbf{0.2415} & \textbf{0.2714} & \textbf{13.82} & \textbf{8.62} & \textbf{2.53} & \textbf{0.41}\\
& 7          & 0.3011 (0.1\%$\downarrow$) & 0.2411 (0.2\%$\downarrow$) & 0.2711 (0.1\%$\downarrow$) & 13.43 (2.9\%$\downarrow$) & 8.31 (3.7\%$\downarrow$) & 2.62 & 0.47\\
& 10         & 0.3008 (0.2\%$\downarrow$) & 0.2412 (0.1\%$\downarrow$) & 0.2710 (0.1\%$\downarrow$) & 12.89 (7.2\%$\downarrow$) & 7.93 (8.7\%$\downarrow$) & 2.76 & 0.56 \\
\hline
\end{tabular}
}
}
\label{tab:number_of_sources}
\end{table*}

\begin{table}[t]
\centering
\caption{Effect of the MDR iteration setting $K$ on reconstruction accuracy and efficiency on the DTU dataset.}
\begin{tabular}{c c c c c c}
\toprule
$K$ & Acc. $\downarrow$ & Comp. $\downarrow$ & Overall $\downarrow$ & Mem. $\downarrow$ & Time $\downarrow$ \\
\midrule
{[1,1,1,1]}   & 0.3031 & 0.2465 & 0.2748 & 2.53 & 0.25 \\
{[3,3,3,3]}   & 0.3027 & 0.2459 & 0.2743 & 2.53 & 0.31 \\
{[12,8,8,3]}  & 0.3011 & 0.2413 & 0.2712 & 2.53 & 0.44 \\
{[12,8,5,5]}  & 0.3009 & 0.2407 & 0.2708 & 2.53 & 0.46 \\
{[12,8,5,3]}  & 0.3013 & 0.2415 & 0.2714 & 2.53 & 0.41 \\
\bottomrule
\end{tabular}
\label{tab:K_iteration}
\end{table}

\noindent \textbf{Effect of MDR Iteration.} Table~\ref{tab:K_iteration} evaluates the impact of the number of MDR iterations $K$ on the DTU dataset. Overall, increasing the iteration count enhances reconstruction quality, particularly in completeness and overall error. This confirms that iterative mutual refinement effectively resolves local scale inconsistencies and recovers missing structures. However, these gains present a clear accuracy-efficiency trade-off. For instance, the low-iteration setting ([3, 3, 3, 3]) significantly reduces computation time from 0.41s to 0.31s, with a marginal increase in overall error from 0.2714 to 0.2743. These findings highlight the practical flexibility of MDR: higher iterations maximize reconstruction accuracy, whereas lower iterations provide substantial speedups with minimal performance degradation, allowing the framework to easily adapt to varying computational constraints.

\subsection{Additional Qualitative Results}

\subsubsection{Qualitative Results on Sparse-View Reconstruction}

\begin{figure}[t]
    \centering
    \includegraphics[width=\linewidth]{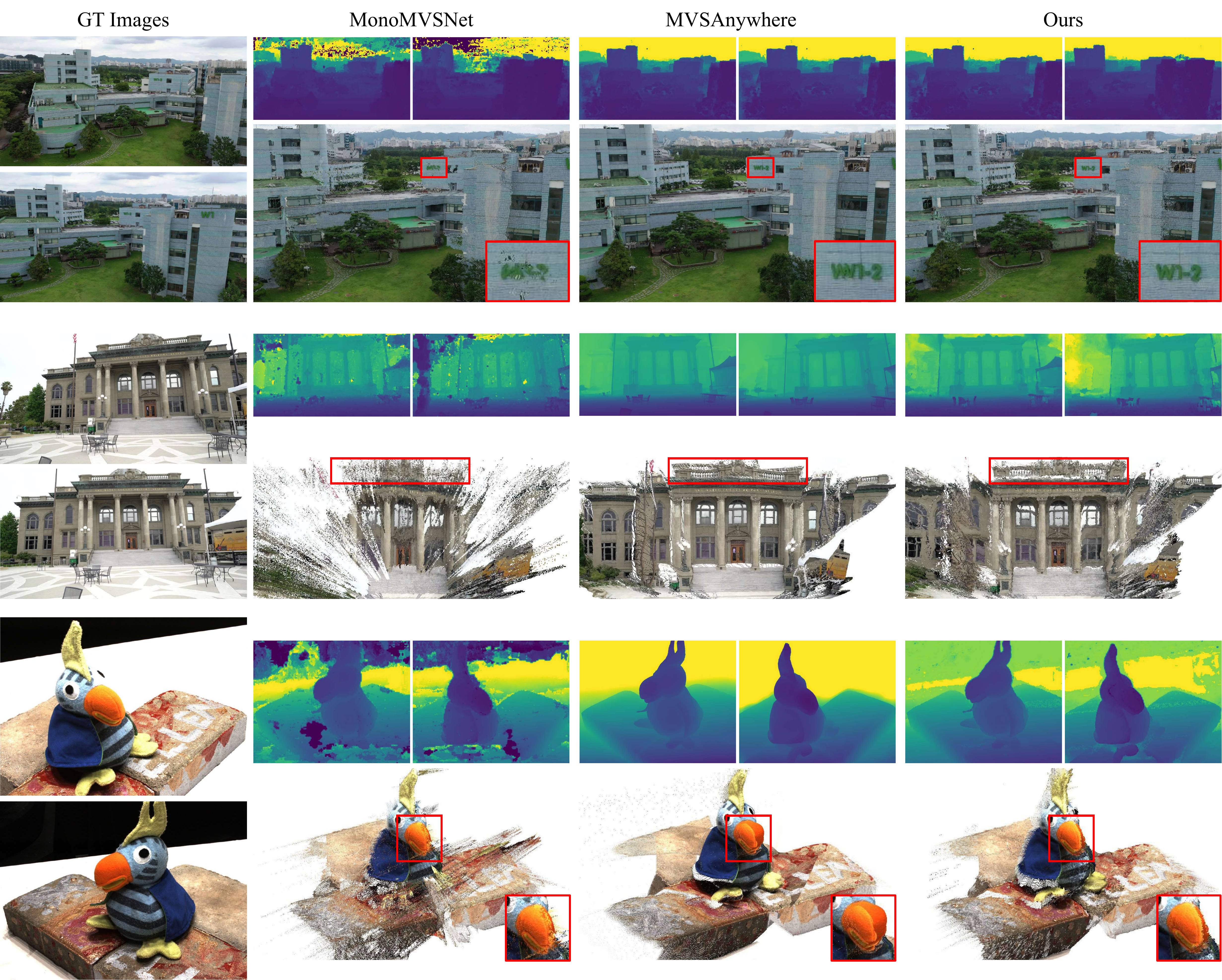}
    \caption{Qualitative comparison of sparse-view 3D reconstruction using only two input images. The top row for each scene displays the estimated depth maps, and the bottom row shows the fused point clouds. MonoMVSNet~\cite{jiang2025monomvsnet} fails to find reliable correspondences due to limited overlap, resulting in severe noise and missing structures. MVSAnywhere~\cite{izquierdo2025mvsanywhere} generates smooth depth maps but suffers from scale ambiguity, leading to misaligned geometries. In contrast, Ours produces highly complete and geometrically accurate reconstructions.}
    \label{sparse_3_views}
\end{figure}

To evaluate the robustness of our method under extreme geometric constraints, we conducted a 3D reconstruction experiment using only two input views. Fig.~\ref{sparse_3_views} presents the qualitative comparison against state-of-the-art baselines.
As shown in Fig.~\ref{sparse_3_views}, MonoMVSNet fails to establish reliable correspondences due to the lack of view overlap, resulting in severe noise and incomplete structures (e.g., the scattered point cloud of the building in the second row). MVSAnywhere, while generating visually smooth depth maps via monocular priors, suffers from scale ambiguity and geometric misalignment.

Our method successfully unifies the structural completeness of monocular priors with the metric accuracy of multi-view stereo. Even with minimal multi-view cues, our method generates geometrically consistent and highly detailed reconstructions.

\subsubsection{Qualitative Results on DTU}

Fig.~\ref{fig:qualitative_depth_all}(a) presents an extensive qualitative comparison of depth maps estimated by different methods on the DTU dataset. We compare our method against the state-of-the-art MVS baseline (MVSFormer++), a prior-guided method (MonoMVSNet), and the outputs of the Depth Foundation Model (DFM). Here, $\text{DFM}_{\text{init}}$ denotes the initial monocular depth map predicted by the DFM, while $\text{DFM}_{\text{refined}}$ represents the monocular depth map scale-corrected via our MDR process.

As shown in Fig.~\ref{fig:qualitative_depth_all}(a), MVSFormer++ often fails in textureless or non-Lambertian regions (e.g., the box surface in the 5th row and the fruit in the 3rd row), resulting in incomplete depth maps with significant holes. MonoMVSNet, despite incorporating priors, tends to generate noisy artifacts and jagged boundaries rather than smooth surfaces. Regarding the monocular priors, it can be observed that while $\text{DFM}_{\text{init}}$ exhibits clean depth maps with sharp object boundaries, it suffers from significant scale misalignment compared to the GT. In contrast, $\text{DFM}_{\text{refined}}$ demonstrates substantial improvement, effectively correcting this scale ambiguity while preserving structural details.

Ours successfully integrates the structural completeness of the monocular prior with the geometric accuracy of MVS. It effectively fills the holes present in MVSFormer++ and suppresses the noise observed in MonoMVSNet. Notably, our result closely resembles $\text{DFM}_{\text{refined}}$ but is metrically refined through our mutual refinement process, yielding high-fidelity depth maps that accurately capture fine details and sharp object boundaries comparable to the GT.

\subsubsection{Qualitative Results on Tanks and Temples}

To demonstrate the generalization capability of our method, we provide comprehensive qualitative evaluations on the Tanks and Temples benchmark. Fig.~\ref{fig:qualitative_depth_all}(b) presents a visual comparison of the estimated depth maps. MVS methods often falter in non-Lambertian or textureless regions. As shown in Fig.~\ref{fig:qualitative_depth_all}(b), MVSFormer++ fails to estimate valid depths for reflective surfaces (e.g., the windows in the 4th row) and textureless backgrounds (e.g., the sky in the 1st row), leading to significant holes. MonoMVSNet attempts to densify the result but suffers from severe noise and artifacts. In contrast, Ours effectively leverages the structural prior from the DFM. Notably, by aligning the scale of the monocular prior via our mutual refinement (visualized as $\text{DFM}_{\text{refined}}$), our method produces depth maps that are not only structurally complete but also geometrically consistent, preserving sharp boundaries even in challenging indoor and outdoor scenes.

\subsubsection{Qualitative Results on KITTI Dataset}

\begin{figure}[t]
    \centering

    \begin{subfigure}{\linewidth}
        \centering
        \includegraphics[width=\linewidth]{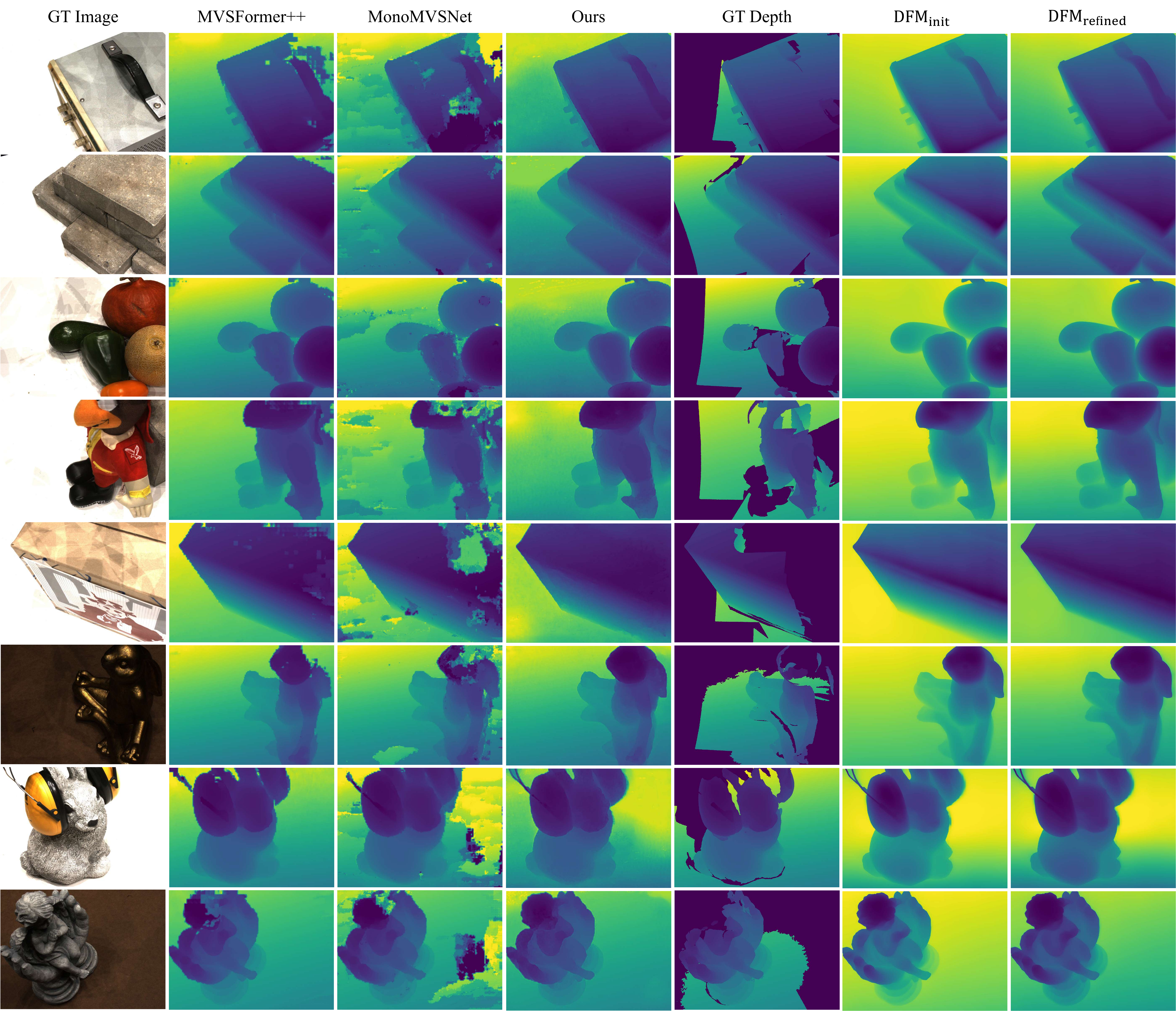}
        \caption{DTU dataset}
        \label{fig:dtu_full_depth}
    \end{subfigure}

    \vspace{0.5em}

    \begin{subfigure}{\linewidth}
        \centering
        \includegraphics[width=\linewidth]{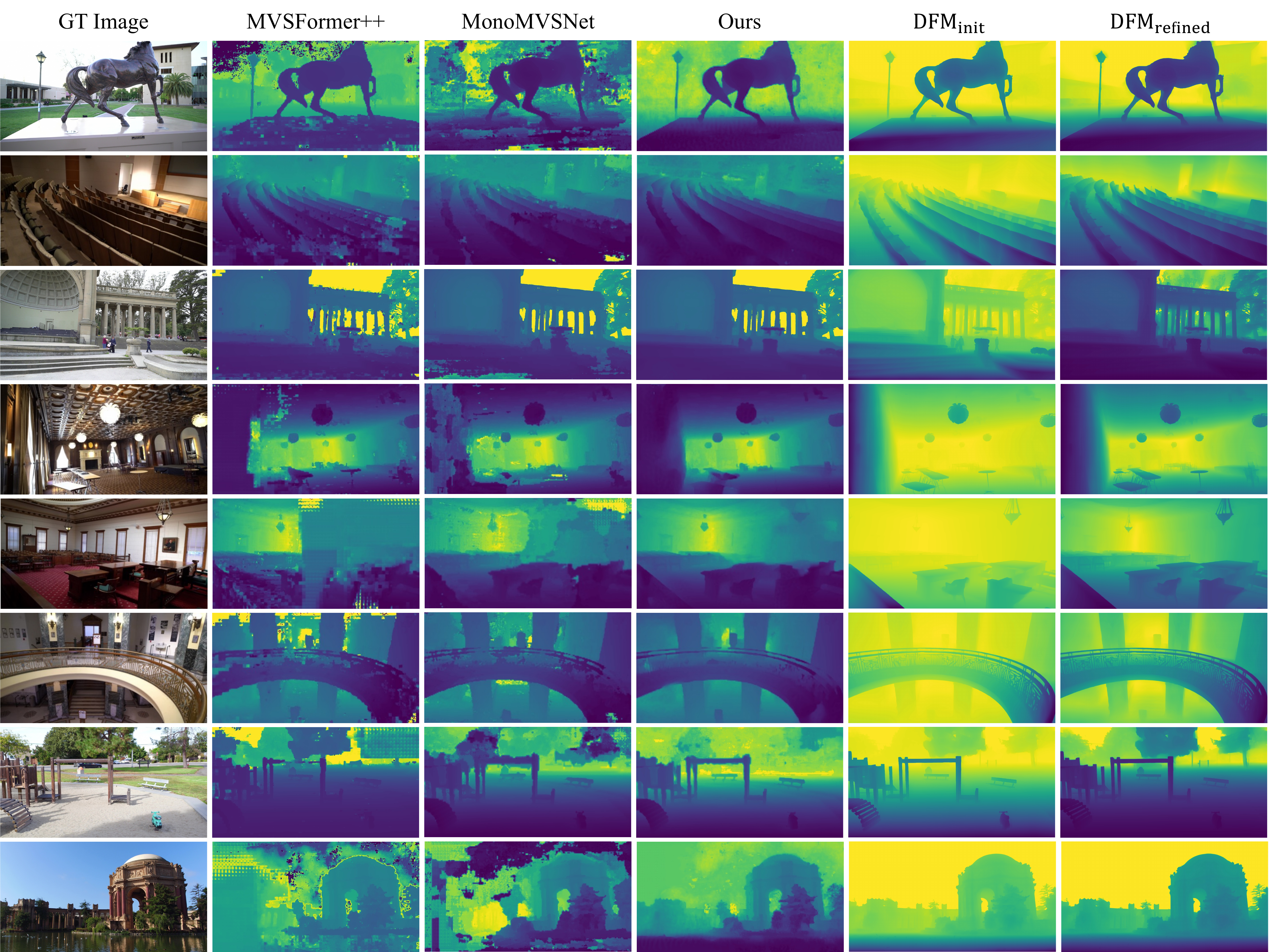}
        \caption{Tanks and Temples dataset}
        \label{fig:tnt_full_depth}
    \end{subfigure}

    \vspace{0.5em}

    \begin{subfigure}{\linewidth}
        \centering
        \includegraphics[width=\linewidth]{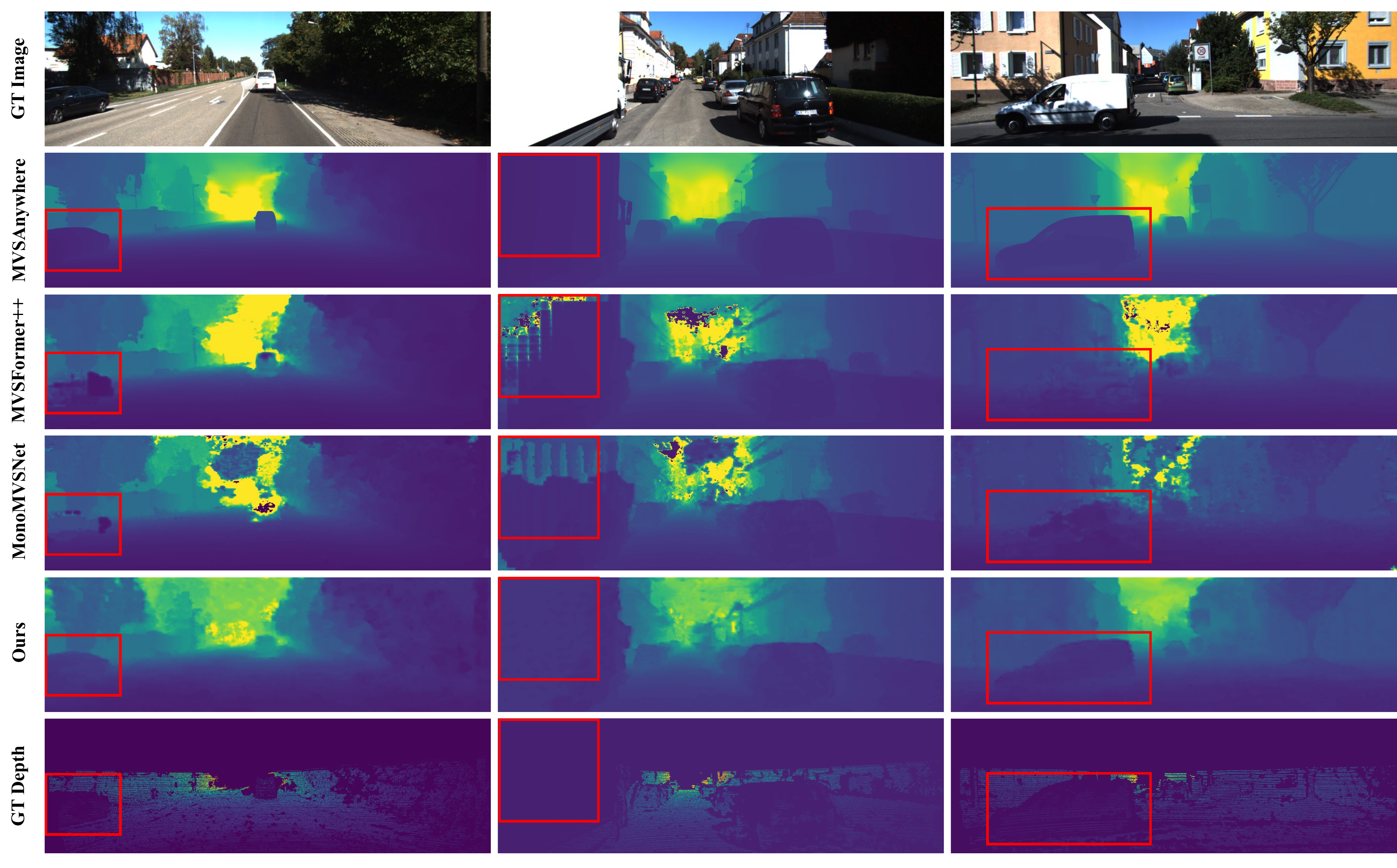}
        \caption{KITTI dataset}
        \label{fig:kitti_depth}
    \end{subfigure}

    \caption{Qualitative comparison of depth maps across three datasets: (a) DTU, (b) Tanks and Temples, and (c) KITTI from RobustMVD.}
    \label{fig:qualitative_depth_all}
\end{figure}

To further assess the generalization capability of our method in outdoor driving scenarios, we conducted a qualitative evaluation using the KITTI dataset provided by the RobustMVD benchmark. Fig.~\ref{fig:qualitative_depth_all}(c) presents a visual comparison against state-of-the-art baselines.

Driving scenes pose unique challenges for MVS, notably due to the presence of dynamic objects (e.g., moving cars) and large textureless regions (e.g., sky and roads). As shown in Fig.~\ref{fig:qualitative_depth_all}(c), MVS-centric methods such as MVSFormer++ and MonoMVSNet  struggle significantly in these areas. They fail to establish reliable correspondences for moving vehicles, resulting in severe noise, artifacts, and incomplete depth estimates. MVSAnywhere, leveraging a DFM-based decoder, generates visually smooth and complete depth maps. However, being a DFM-centric approach, it suffers from inherent scale ambiguity. While the depth map appears clean locally, the absolute depth values often deviate from the metric GT, leading to misaligned geometry in 3D space.

In contrast, our method effectively bridges the gap between these two paradigms. By utilizing the Cost-Crafter to anchor the geometry and MDR to refine structural details, we generate depth maps that are both sharp and scale-consistent. As shown in the red boxes, while our boundary recovery is slightly less sharp compared to MVSAnywhere, our model successfully reconstructs the shapes of dynamic vehicles and preserves smooth gradients on road surfaces. This demonstrates our method's superior robustness compared to both pure MVS and other hybrid approaches.

\bibliographystyle{IEEEtran}
\bibliography{main}

\end{document}